\documentclass[sigconf,,nonacm]{acmart}
\usepackage{xspace}
\usepackage{caption}
\usepackage{subcaption}
\usepackage{float}
\usepackage{placeins}

\makeatletter
\DeclareRobustCommand\onedot{\futurelet\@let@token\@onedot}
\def\@onedot{\ifx\@let@token.\else.\null\fi\xspace}

\def\eg{\emph{e.g}\onedot} 
\def\ie{\emph{i.e}\onedot}

\def\etal{\emph{et al}\onedot}
\makeatother
\usepackage{listings,chngcntr,float}
\definecolor{forestgreen}{rgb}{0.0, 0.27, 0.13}
\definecolor{bluegray}{rgb}{0.4, 0.6, 0.8}
\lstdefinestyle{DefaultStyle}{keywordstyle={\color{forestgreen}\textbf},stringstyle={\color{bluegray}},morekeywords={PWLCurve,EnumCurve,pwlfit,pwlf,fx,identity,sgn,log,log1p,symlog1p,symmetriclog1p,abs,num_samples,mono,num_segments,max_slope,min_slope},
basicstyle=\ttfamily\footnotesize,float}
\AtBeginDocument{%
  \providecommand\BibTeX{{%
    \normalfont B\kern-0.5em{\scshape i\kern-0.25em b}\kern-0.8em\TeX}}}

\begin{document}

\title{Distilling Interpretable Models into Human-Readable Code}

\author[Ravina \etal]{Walker Ravina$^*$, Ethan Sterling, Olexiy Oryeshko, Nathan Bell, Honglei Zhuang, Xuanhui Wang, Yonghui Wu, Alexander Grushetsky}
\thanks{$^*$Corresponding author}
\affiliation{%
  \institution{Google, Mountain View, CA, USA}
}
\email{{walkerravina, esterling, olexiy, nathanbell, hlz, xuanhui, yonghui, grushetsky}@google.com}

\renewcommand{\shortauthors}{Ravina et al.}

\begin{abstract}
The goal of model distillation is to faithfully transfer teacher model knowledge to a model which is faster, more generalizable, more interpretable, or possesses other desirable characteristics. Human-readability is an important and desirable standard for machine-learned model interpretability. Readable models are transparent and can be reviewed, manipulated, and deployed like traditional source code. As a result, such models can be improved outside the context of machine learning and manually edited if desired.  Given that directly training such models is difficult, we propose to train interpretable models using conventional methods, and then distill them into concise, human-readable code.

The proposed distillation methodology approximates a model's univariate numerical functions with piecewise-linear curves in a localized manner. The resulting curve model representations are accurate, concise, human-readable, and well-regularized by construction.  We describe a piece\-wise-linear curve-fitting algorithm that produces high-quality results efficiently and reliably across a broad range of use cases. We demonstrate the effectiveness of the overall distillation technique and our curve-fitting algorithm using four datasets across the tasks of classification, regression, and ranking.
\end{abstract}

%%
%% The code below is copy pasted from the tool at http://dl.acm.org/ccs.cfm.
%%



\maketitle

\section{Introduction}

Interpretable models are critical for high-stakes decision-making scenarios \cite{rudin2018} such as guiding bail or parole decisions, assessing loan eligibility, and guiding medical treatment decisions. In these cases, the explanation of a model's output (\eg{ individual feature contributions}) should be examinable and understandable, to ensure transparency, accountability, and fairness of the outcomes.

To achieve intrinsic interpretability, univariate functions are widely used in interpretable models. In the classic Generalized Additive Models (GAMs) \cite{hastie1986}, the model is a sum of univariate shape functions,
$$
M = f_0 + f_1(x_1) + f_2(x_2) + f_3(x_3) + \dots + f_n(x_n).
$$
where $x_i$'s are $n$ features and $f_i$'s are the shape functions. Such a model is simple but often less accurate than a model with feature interactions. Recently, Lou \etal \cite{lou2013} showed that adding a limited number of pairwise feature interactions allows GAM-style additive models to capture a significant fraction of the accuracy of a fully-interacting model. In many cases of interest, such feature interactions are intuitively captured with products of univariate functions,
$$
g_1(c_1) \cdot f_1(x_1) + g_2(c_2) \cdot f_2(x_2) + \dots,
$$
or products of groups of features,
$$
(g_{1,1}(c_1) + g_{1,2}(c_2)) \cdot f_1(x_1) + (g_{2,1}(c_1) + g_{2,2}(c_2)) \cdot f_2(x_2) + \dots,
$$
where the magnitude of one function (\ie{ $f_i$}) is modulated by a function (\ie{ $g_i$ or $g_{i,j}$}) of another "context" feature (\ie{ $c_i$})~\cite{ngam}. In other cases, the interaction amongst features is adequately approximated by additive models of univariate functions nested within univariate functions,
\begin{align*}
 f(x_1, x_2, x_3) \approx& \ g_1(f_{1,1}(x_1) + f_{1,2}(x_2) + f_{1,3}(x_3)) \ + \\ 
  & \ g_2(f_{2,1}(x_1) + f_{2,2}(x_2) + f_{2,3}(x_3)) + \dots,
\end{align*}
where the outer function $g_i$ captures nonlinear behavior \cite{chen2018}.  Indeed, the Kolmogorov–Arnold representation theorem~\cite{kolmogorov,wiki_Kolmogorov} guarantees that every continuous multivariate function of $n$ inputs can be represented as a sum of $2n$ such terms,
$$
f(x_{1},\dots ,x_{n}) = \sum _{i=0}^{2n} g_{i}\left(\sum _{j=1}^{n}f_{i,j}(x_{j})\right).
$$
In practice a single outer function is often sufficient, yielding an interpretable model.

\begin{figure*}[t]
  \centering
  \includegraphics[width=\textwidth]{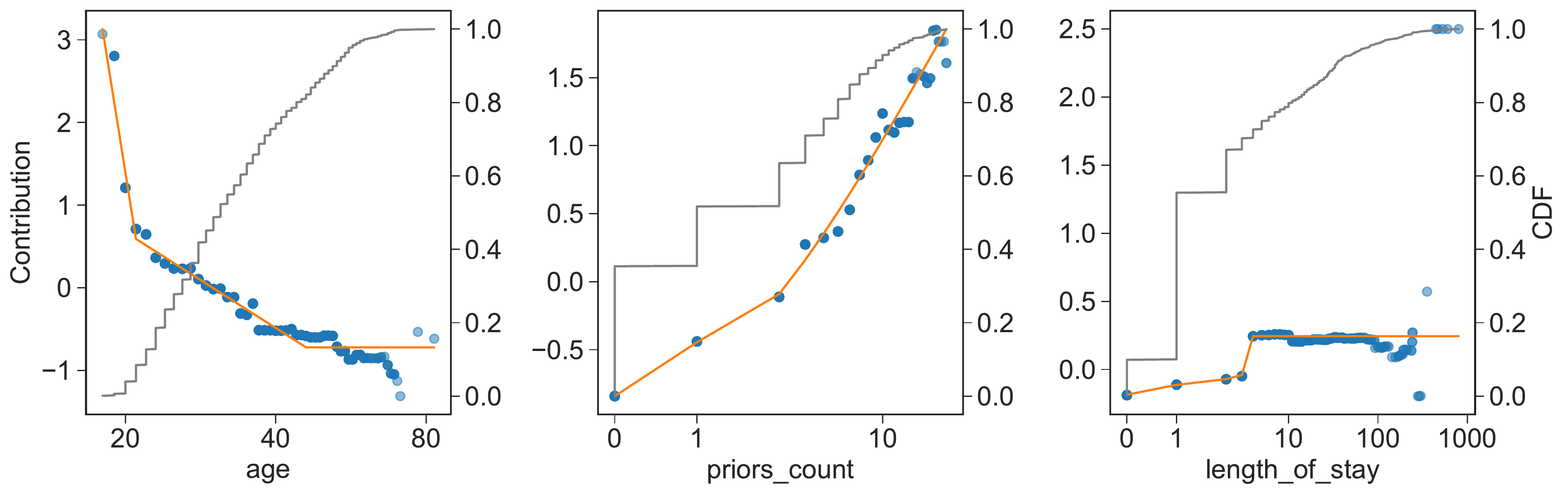}
  \caption{Shape plots for numerical features from models learned on the COMPAS dataset. The GAM forest model is shown in blue dots, while its distillation into the two-segment curve model is shown in orange lines (both map to the left Y axis). Cumulative distribution functions of the corresponding signals are shown in grey (right Y axis).}
  \label{fig:COMPAS_Shape_Plots}
\end{figure*}

In the classic GAM models, splines are used as shape functions~\cite{hastie1986}. Another commonly used shape function is piecewise-linear functions~\cite{wiki_Segmented_regression}. These representations contain a small number of variables (\eg{ knots}) and thus are concise and human-readable. However, directly optimizing such representations often yields less accurate models than alternative model representations. For example, Lou \etal \cite{lou2012} showed that learning spline GAMs is less accurate than learning bagged boosted decision forest GAMs. Our experiments show similar results for directly optimizing GAMs composed of piecewise-linear curves using Stochastic Gradient Descent (SGD) methods. Broadly speaking, the model representations using decision forest GAMs have the advantage during model optimization, but the resultant models are not human-readable. This is the case even when there exists a simpler model with a concise, human-readable form that provides comparable accuracy. 

Inspired by the model distillation work in which relatively small decision forests or neural networks can be distilled from much larger ensembles, but not trained directly from data, to match the accuracy of complex models \cite{buciluundefined2006,hinton2015distilling}, we propose to distill interpretable models into readable representations in a separate process after model optimization. This decouples the initial, learned model representation from the final, published model representation. For example, the proposed distillation methodology can be applied to additive models trained using bagged boosted decision trees~\cite{lou2012}, as well as additive neural nets \cite{nam,ngam}. 

In this paper, we describe a technique for distilling models composed of univariate components into human readable representations, in particular, the piecewise-linear curves described in Section~\ref{sec:pwlcurve}.
The output of our distillation technique is illustrated in Listing~\ref{lst:COMPAS_code} and Figure~\ref{fig:COMPAS_Shape_Plots}, which show textual and graphical representations of piecewise-linear curves obtained by applying our approach to a decision forest GAM trained on the COMPAS dataset (described in Section~\ref{sec:data-description}). The distilled model is a concise representation of the decision forest GAM model and is converted to human-readable source code.
\begin{lstlisting}[caption=Code for a distilled COMPAS model,label={lst:COMPAS_code},frame=single,language=Python,style=DefaultStyle,breaklines=true]
score = sum([ 
  PWLCurve("age", [(18, 3.13), (21, 0.5914), 
    (46, -0.7206)], fx="log"),
  PWLCurve("priors_count", [(0, -0.8415), (1, -0.4452), (38, 2.146)], fx="log1p"),
  PWLCurve("length_of_stay", [(0, -0.1855), 
    (3, -0.04099), (4, 0.2443)], fx="log1p"),
  EnumCurve("c_charge_degree", {1: 0.0198, 2: -0.0384}),
  ## ... other features ...
])
\end{lstlisting}
From here on, we will use "curves" to refer to piecewise-linear curves, "curve models" to refer to models where each component is a curve, and "code" to refer to the textual representations of curve models or curves. 

The rest of this paper is structured as follows. After presenting the preliminaries in Section~\ref{sec:preliminaries}, we elaborate on the benefits of using curve models in Section~\ref{sec:motivation}. We then describe the localized distillation process in Section~\ref{sec:Localized Distillation} and piecewise-linear approximation algorithm, sometimes referred to as segmented regression \cite{wiki_Segmented_regression}, for creating curve models in Section~\ref{sec:Piecewise-Linear Curve Approximation}. Lastly, we present experimental results on for datasets: COMPAS, FICO, MSLR-WEB30K, and CWS in Section~\ref{sec:exp} and conclude the paper in Section~\ref{sec:conclusion}.

\section{Preliminaries}\label{sec:preliminaries}
Throughout the paper, we will use the data sets used in this paper as concrete examples to explain our methods. Thus, we first describe them in this section. We also give the formal definition of piecewise-linear-curves in this section.

\subsection{Data Sets}\label{sec:data-description}
We used the following four datasets to represent different settings: classification, regression, and ranking. The first three are publicly available.
\begin{itemize}
    \item The COMPAS dataset\footnote{\url{https://github.com/propublica/compas-analysis}} is the result of a ProPublica investigation \cite{angwin2016machine} into possible racial bias of the proprietary COMPAS model score for defendants in Broward county, Florida. The dataset has been studied extensively in the context of bias, fairness, and interpretability \cite{tan2018, dressel2018, kleinberg2016, chouldechova2017}. Labels are binary and indicate whether recidivism occurred for an individual within a time period. We use area under the receiver operating characteristic curve (AUC-ROC) to measure classifier accuracy. COMPAS has 6 features and four of them are used as examples in this paper: \lstinline{age}, \lstinline{priors_count}, \lstinline{length_of_stay}, and \lstinline{c_charge_degree}.
    \item The FICO dataset \cite{FICO} is composed of real-world anonymized credit applications along with risk scores. Labels are a risk score for an individual. We use root mean square error (RMSE) to measure regressor accuracy. FICO has 24 features and we use two features as examples in our paper:  \lstinline{MSinceMostRecentDelq}, Months Since Most Recent Delinquency; \lstinline{PercentTradesWBalance}, Percent Trades with Balance. 
    \item The MSLR-WEB30K dataset \cite{mslr} is a widely used learning-to-rank benchmark dataset. Labels are per document relevance judgements. We use normalized discounted cumulative gain at $k=5$ (NDCG@5) to measure ranker accuracy. MSLR-WEB30K is significantly larger both in number of features (136) and number of training examples (\textasciitilde{2,000,000} per cross validation fold). We use it to compare our curve approximation algorithm to \lstinline{pwlf}  \cite{pwlf}, a publicly available alternative, on the basis of accuracy, robustness and efficiency. We use two features as examples in our paper: \lstinline{feature_0011}, Body stream length; \lstinline{feature_0128}, Inlink number.
    \item The Chrome Web Store (CWS) dataset is a private and anony\-mized dataset originating from Chrome Web Store logs. Each query corresponds to a visit to the Chrome Web Store. The items within each query were the ones shown to the user. Labels correspond to user actions such as clicking, installing, or no action whatsoever. We again use NDCG@5 to measure ranker accuracy. A similar, but distinct datset from the Chrome Web Store was previously studied by Zhaung \etal \cite{ngam}. Unlike in that previous work, in this instance we do not utilize query level "context" features, instead using only 14 item level features. The queries are also distinct.
\end{itemize}
In each case, we distill a decision forest GAM and evaluate the accuracy of the distilled curve models. The COMPAS and FICO datasets represent high-stakes domains \cite{rudin2018} in which the benefits of curve models, discussed below, are particularly compelling. FICO, MSLR-WEB30K, and CWS have been previously studied in the context of interpretability  \cite{nam, ngam, lou2013, chen2018}. Furthermore, the results from MSLR-WEB30K demonstrate that the accuracy of this approach is not limited to small datasets.

\subsection{Piecewise-Linear Curves}\label{sec:pwlcurve}
A piecewise linear curve (\lstinline{PWLCurve}) is defined by a list of control points $S=[(x_k, y_k)]_{k=1}^K$ through which the curve must pass. Between control points, output $y$ values are determined by performing linear interpolation between neighboring control points. Beyond the leftmost or rightmost control points, output values are capped to the $y_k$-value of the neighboring control point.  More formally,
assuming $x_k$'s are ordered,~\ie~$x_k < x_{k+1}$,
the definition of a piecewise linear curve can be described as:
\begin{equation*}
PWL(x;S) = 
    \begin{cases}
        y_1 & \text{if } x < x_1, \\
        \frac{y_{k+1} - y_k}{x_{k+1} - x_k} (x - x_k) + y_k & \text{if } x_k \leq x \leq x_{k+1}, \\
        y_K & \text{if } x > x_K.
    \end{cases}
\end{equation*}
In most cases of interest 5 or 6 control points, defining 4 or 5 interior segments, is sufficient to capture the desired behavior. 

We allow for an optional $x$-transformation, specified with the \lstinline{fx} argument, to fit curves to data with different scales. When an $x$-transformation is present it is applied to the input value and $x$-values of all the control points, and then linear interpolation is performed in the transformed space. We support \lstinline{identity} (default), \lstinline{log}, \lstinline{log1p} and \lstinline{symlog1p} transformations.  Here \lstinline{symlog1p} is defined as \lstinline{sgn(x) * log1p(abs(x))} and is suitable for highly-variable features that take on both positive and negative values.

Univariate categorical functions are represented by \lstinline{EnumCurve}, which directly maps input values to outputs using a discrete mapping.

\section{Background \& Motivation}\label{sec:motivation}
Interpretable models are critical for high-stakes decisions \cite{rudin2018} and provide many advantages over more complex model structures \cite{caruana2015, du2019}.  In this section we explain how distilling interpretable models into curve models reinforces these benefits and addresses a variety of real-world engineering challenges. Here, one underlying theme is that distilling models into human-readable source code \emph{reduces a novel machine learning problem to an established software engineering problem with an abundance of existing solutions}.

\subsection{Greater Transparency}
\label{subsec:greater-transparency}
\begin{figure}
  \centering
  \includegraphics[width=\linewidth]{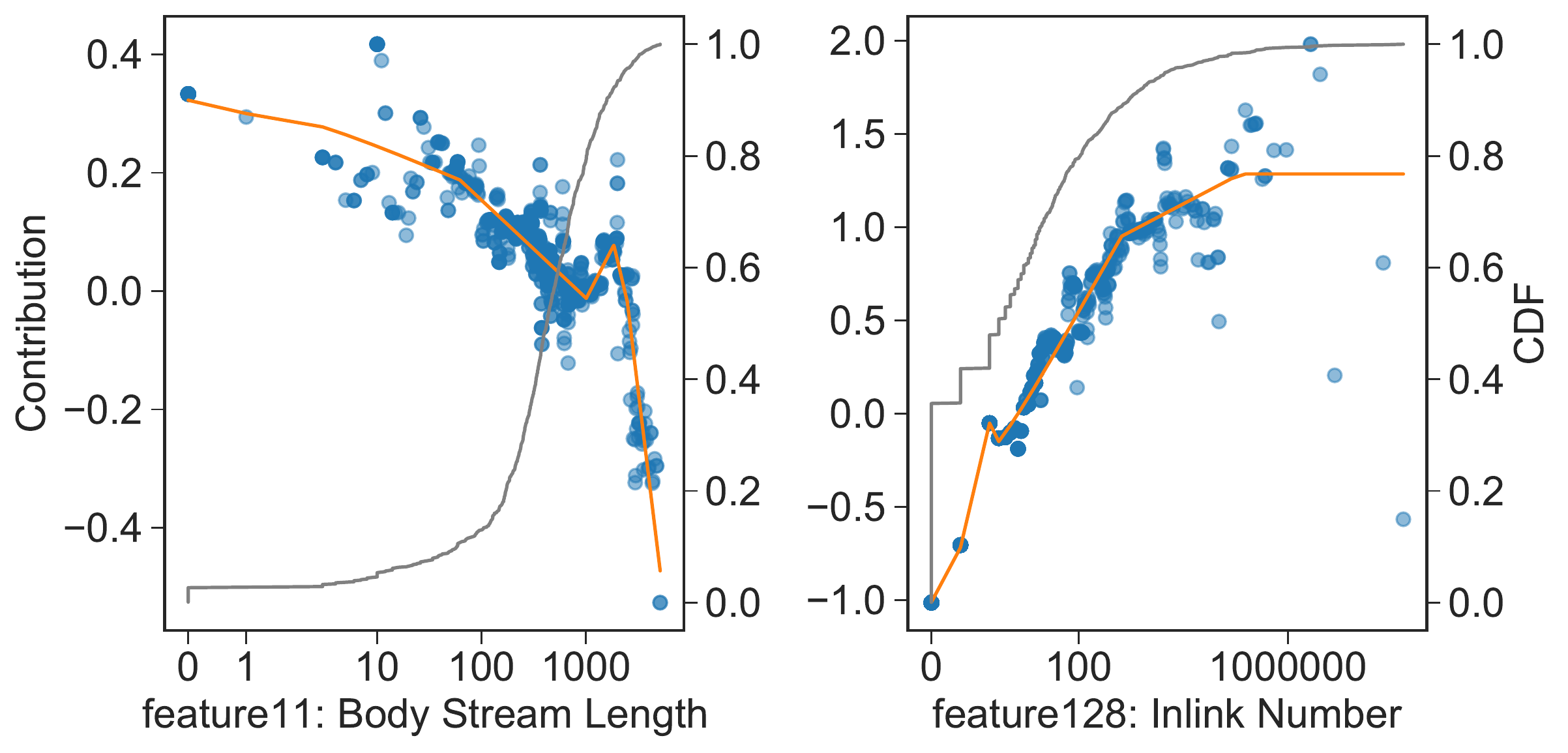}
  \caption{Shape plots for a GAM forest model (in blue dots) and 5 segment curve distillation (in orange lines) for the MSLR-WEB30K dataset.} 
  \label{fig:MSLR_regularize}
\end{figure}

A model is transparent if it provides a textual or graphical representation that enables its behavior to be understood comprehensively \cite{ustun2014}.  One way in which the proposed method provides greater transparency is by simplifying graphical depictions of a model while retaining its essential characteristics. It is often argued, implicitly or explicitly, that the shape plots of an interpretable model are an \emph{exact description} of the model and therefore provide a reliable way to understand the model.  While this claim is narrowly true, it is misleading in general.  Unless given specific guidance, humans will naturally discount certain fine-grained details of the plots when developing an understanding of the model.  By distilling interpretable models to a concise representation, we discard extraneous characteristics and reduce the mental effort necessary to understand the model.  For example, it is not immediately obvious what understanding an individual should derive from the shape plots of the \lstinline{feature_0011} (body stream length), and \lstinline{feature_0128} (inlink number) features in the initially-learned MSLR-WEB30K model, shown in Figure~\ref{fig:MSLR_regularize}. Indeed, different individuals may derive qualitatively different understandings from these graphical depictions.  However, given the additional knowledge that the distilled curve model represented by the overlaid curves in Figure~\ref{fig:MSLR_regularize} has nearly identical accuracy, an observer can make much stronger inferences about the model’s essential characteristics. Interpretability can be increased even further by imposing monotonicity constraints. We discuss the effect of such constraints in Section~\ref{subsec:Monotonicity}.

Clearly when distillation yields a simpler model with comparable accuracy we would say the distillation process has succeeded.  However, instances where distillation yields a model with inferior accuracy warrant further investigation because the apparent "failure” can often be attributed to essential characteristics of the teacher model that were not successfully transferred to the student \emph{because they violate a prescribed notion of human-interpretability}. We examine one representative example of this phenomenon in Section~\ref{subsec:distillation-failures}. While a complete discussion of this principle is beyond the scope of this paper, we note that the idea can be viewed as an extension of the use of structural constraints to define “interpretable” models, just now applied to the structure of individual functions in the model.  Under this policy, if the accuracy of a candidate model cannot be reproduced using a predefined class of expressive, “human-scale” functions (\eg{ curves with a small number of truncated control points}) its transparency would be called into question.

\subsection{Constructive Regularization}
\label{subsec:Constructive Regularization}

The proposed method can also be viewed as a post-hoc regularization process that is completely compatible with, and complementary to, optimization-based regularization techniques (\eg{ L1/L2 penalties or monotonicity constraints}).  In the context of regularization, our emphasis on conciseness is aligned with the minimum description length principle \cite{wiki_Minimum_description_length} for model selection.  Ustun and Rudin \cite{ustun2014} applied similar reasoning to motivate linear models with small, integer-valued weights.  The constrained description length of curves provides limited capacity for capturing idiosyncratic behavior. As a result, curve distillation successfully removes aberrations from teacher model functions. This regularization effect can be seen in Figure~\ref{fig:COMPAS_Shape_Plots} and Figure~\ref{fig:MSLR_regularize}. The fewer segments the greater the effect. To find the most concise curve model we can repeatedly apply the proposed method with decreasing number of control points.  Naturally, the optimality of this approach is subject to the limitations of our localized distillation methodology (see Section~\ref{sec:Localized Distillation}) and curve approximation algorithm (see Section~\ref{sec:Piecewise-Linear Curve Approximation}).  While it is difficult to directly compare models with different functional representations, comparing the length and readability of their corresponding code is instructive.

One practical advantage of curve-based regularization is that regularity is enforced by construction and the complexity of individual curves is readily apparent and quantifiable.  Therefore, organizations that adopt curve models can set objective guidelines about model complexity that developers can anticipate when submitting model candidates for approval. Such guidelines can specify the maximum number of curve segments, maximum number of significant digits per curve control point, or monotonicity of the curve. Similar to the use of nothing-up-my-sleeve numbers in cryptography \cite{wiki_Nothing-up-my-sleeve_number}, curve models enable developers to preemptively address suspicions about potential weaknesses and constructively prove the robustness of a given model candidate. In general, standardizing development around curve models is a straightforward way for organizations to systematically enforce best practices, defend against common mistakes and pitfalls, and expedite model verification and approval. The accessible, readable nature of curve models enables organization members beyond engineers (\eg{ executives, product managers, etc.}) to participate in this approval process. 

\subsection{Readable, Editable Code}

Curve model code can be read, reviewed, merged, and versioned like conventional source code. An example model for the COMPAS dataset is shown in Listing~\ref{lst:COMPAS_code}.  One can understand how a curve model would behave under novel or extremal conditions by mentally “evaluating” the model under hypothetical “what if?” scenarios without the need for additional tools.  Subjecting models to a traditional source code review process facilitates a more rigorous examination of the model’s characteristics and greater accountability than is possible with non-readable models.  Indeed, conducting “model review” through source code review ensures that the candidate model itself - not some separate, potentially inconsistent description or artifact of the model or how it was trained - is the subject of review. In the event that undesirable model behavior is discovered, the model's code may be directly edited to correct such issues. For example, in the case of the COMPAS model a user may wish to deliberately cap the contribution of features such as \lstinline{priors_count} and \lstinline{length_of_stay} features for legitimate policy reasons not captured by classification metrics such as AUC-ROC. The contribution of other features can be entirely removed. Agarwal \etal \cite{nam} discussed how such an approach of training with biased features and then removing them can potentially be better than simply training without biased features. This approach can prevent the model from extracting bias through other features which are correlated with biased ones.

Model transparency is essential in the context of high-stakes decisions \cite{rudin2018} arising in criminal justice, finance, health care, and other areas. Providing the complete source of the model in simple, portable, human-readable code makes the models transparent. Compared to human-readable models produced by CORELS \cite{corels}, which are expressed in universally-understandable if-then language, curve models sacrifice accessibility for greater expressiveness and general-purpose application.

\subsection{Collaborative Model Development}

Curve distillation is compatible with any algorithm or modeling technique that results in univariate functions. In the experiments section we apply the proposed technique to decision forest GAMs on several datasets. Previous work \cite{ngam} applied the proposed technique to GAMs learned via neural networks, as well as similar neural networks with limited interactions via multiplicative pairs. Organizing collaborative development around curve models enables engineers to apply a plurality of different tools, techniques, or platforms to optimize components of a (potentially large-scale) model. Engineers are free to choose a modeling approach that maximizes their productivity, similarly to how engineers use multiple IDEs, code formatters, or linters to collaboratively develop software. Curve distillation can be viewed as a “format conversion” tool that translates an arbitrary and potentially exotic model representation into a fixed, agreed-upon vocabulary of human-readable building blocks.

\subsection{Straightforward Deployment}

Curve models are fast-to-evaluate and straightforward to deploy. Since evaluation requires minimal computation - just a handful of floating point operations per curve - curve models are well-suited for performance-critical applications.  Curves are a portable, platform-agnostic representation that can be natively supported in a variety of languages or systems with little effort.  For example, Listing~\ref{lst:COMPAS_cpp} shows a C++ implementation of a COMPAS model with 2 segments.  In general, curve models are straightforward to deploy because they offer a multitude of integration options.  Curves can be embedded in configuration files, passed via CGI parameters, manually embedded into complex applications in a piecemeal fashion, systematically translated to a target representation, or evaluated by existing runtime systems with a few incremental extensions.

\begin{lstlisting}[caption=A COMPAS model as a C++ function,label={lst:COMPAS_cpp},frame=single,language=C++,style=DefaultStyle,breaklines=true]
double COMPAS(double age, double priors_count,
              double length_of_stay, int charge_degree,
              // ... other features ...
              ) {
  static auto age_curve = PWLCurve({{18, 3.13}, 
    {21, 0.5914}, {46, -0.7206}}, "log");
  static auto priors_count_curve = PWLCurve(
    {{0, -0.8415}, {1, -0.4452}, {38, 2.146}},"log1p");
  static auto length_of_stay_curve = PWLCurve(
    {{0, -0.1855}, {3, -0.04099}, {4, 0.2443}}, "log1p");
  static auto charge_degree_curve = EnumCurve({{1, 0.0198}, {2, -0.0384}});
  // ... other features ...
  return (age_curve.Eval(age) + 
    priors_count_curve.Eval(priors_count) + 
    length_of_stay_curve.Eval(length_of_stay) + 
    charge_degree_curve.Eval(charge_degree) + 
    // ... other features ...
  );
}
\end{lstlisting}

\section{Localized Distillation}
\label{sec:Localized Distillation}

Our distillation process takes two inputs: a teacher model containing one or more univariate functions, and a representative dataset (generally the training data).  Our method differs from conventional distillation techniques in that we (1) distill each univariate function in isolation and (2) optimize for mean squared error (MSE) when approximating each univariate function. Specifically, each univariate function in the teacher model is evaluated on the dataset to produce representative $(x, y)$ example pairs. For discrete categorical features we create a mapping where each unique $x$ is mapped to the mean $y$. For numerical features, we produce a \lstinline{PWLCurve} using the approximation algorithm described in Section~\ref{sec:Piecewise-Linear Curve Approximation}. If the teacher model contains univariate functions nested within other univariate functions we replace the source functions in a bottom-up fashion. Otherwise, all non-nested functions can be approximated in parallel. The final model is constructed by replacing each original univariate function with its \lstinline{PWLCurve} approximation. 

Conventionally, model distillation involves a global optimization using the same (or at least similar) objective to the original teacher model training. This objective may differ from a point-wise MSE objective. For example, ranking objectives often have pair-wise definitions.  Why then do we advocate a localized optimization using a MSE objective in all circumstances?  The primary answer is that, in the context of interpretable models, there is substantial value in maintaining a strong one-for-one correspondence between each source function and target function.  Notably, this allows us to visualize each shape function in the teacher model against its corresponding curve replacement.  Additionally, we can attribute distillation failures - data instances where the curve model is less accurate than the teacher model - to specific univariate functions, and to take remedial actions.  For example, in the Figure~\ref{fig:cws_bad_fit} we can immediately tell that the shape function of $x_1$ was not well-approximated by a curve. In the experiments section we show that the meaningful behavior of nearly all shape functions can be accurately captured by curves with three to five segments. Furthermore, when the meaningful behavior is not captured, it is generally due to inherently non-interpretable behavior being lost. 

While a global optimization approach (\ie{ optimizing the parameters of all curves in the target model simultaneously}) using a problem-specific metric might produce a more accurate result, it is computationally more expensive and would lack the same one-to-one correspondence with the teacher model, making distillation failures more difficult to diagnose.  Furthermore, if higher accuracy is desired, the output of the proposed distillation process can be used to initialize a global optimization of the curve model’s parameters.

\section{Piecewise-Linear Curve Approximation}
\label{sec:Piecewise-Linear Curve Approximation}

Given a univariate numerical function $f(x) \rightarrow y$, our goal is to produce a \lstinline{PWLCurve}  $c(x) \rightarrow y$ that faithfully approximates $f(x)$ by minimizing the $MSE(c(x), f(x))$ over sample data.  Clearly, the accuracy of the overall distillation method depends critically on the accuracy of the individual curve approximations - \ie{ how much metric loss is incurred when each $c(x)$ is substituted for the corresponding $f(x)$ in the trained model}.

Additionally, the practical success of the methodology also depends on the robustness and efficiency of the approximation algorithm.  To enable systematic use of curve distillation in model training pipelines, the approximation algorithm must run with minimal configuration. Complex hyperparameters pose a significant barrier to entry. We have designed \lstinline{pwlfit}, our piecewise linear approximation algorithm, so that in practice users only need to consider the \lstinline{num_segments} and \lstinline{mono} (monotonicity)  parameters. While \lstinline{num_segments=5}  segments and \lstinline{mono=False} is sufficient to get high accuracy (as demonstrated by our experiments), it is desirable to investigate whether the model can be further simplified with fewer segments or with monotonicity restrictions. To facilitate such investigations it is important that distillation runs quickly (\eg{ less than 1 second per function}) which enables interactive analysis via Jupyter notebooks \cite{jupyter} or other tools. These practical considerations have informed various decisions in the design of \lstinline{pwlfit}. In particular, we prefer an algorithm which quickly and reliably yields high accuracy results with minimal configuration to one which sacrifices either of these practical considerations for marginal gains in accuracy. 

In this section we will describe the salient characteristics and noteworthy features of \lstinline{pwlfit}.  We invite interested readers to consult the publicly-available source code of \lstinline{pwlfit} \cite{pwlfit}, for additional details.

\subsection{Algorithm}

Given a list of $(x, y, weight)$ points and a desired number of segments $k$, we search for a \lstinline{PWLCurve} to minimize mean squared error, MSE. A \lstinline{PWLCurve} with $k$ segments is characterized by its $k + 1$ control points -- a set of $x$-knots and their corresponding $y$-knots. Given only the $x$-knots, we can solve a linear least squares expression for the optimal $y$-knots and the resulting error. Since we don't know the correct $x$-knots, we search through the space of possible $x$-knots and solve a least squares expression at each step to calculate the error.\footnote{\lstinline{pwlf} \cite{pwlf} implements a similar approach. We will compare with it in our experiments.}

\subsubsection{Initial Downsampling}

For performance, we randomly downsample large datasets to approximately one million points before fitting. We downsample to reduce the cost of sorting, which dominates the runtime for large data. This downsampling imposes a negligible quality loss.
To further reduce runtime, we discretize the search space for $x$-knots. We choose \lstinline{num_samples} $x$-values from the data, spaced equally by cumulative weight, and search over the combinations of $x$-knots from that sampled set of candidates. Using the default 100 samples, our candidates are the $x$-values at $(0\%, 1.01\%,\dots, 98.9\%, 100\%)$ of the cumulative weight.

\subsubsection{Knot Discretization}

For data with many repeated $x$-values, some of our candidates will be duplicates. For example, $55\%$ of the values in the \lstinline{length_of_stay} feature in the COMPAS data set are 0 or 1. In such cases, we iteratively resample at higher rates (such as $0\%, 0.505\%, 1.01\%$, etc.) until we collect a suitable number of distinct candidates, never exceeding the specified \lstinline{num_samples} parameter.

\subsubsection{Condensation}

To minimize the cost of each linear least squares step, we condense the data using a novel technique described in Appendix~\ref{appendix:linear_condense}. Given \lstinline{num_samples} candidate knots, we condense the full data into two synthetic points per adjacent pair of candidates, for a total of \lstinline{2 * (num_samples - 1)}  synthetic points. For any function that's linear between each adjacent pair of candidate $x$-knots, which is guaranteed by our choice of discrete candidate $x$-knots, these condensed points perfectly recreate the loss of that function over the full data set. We run our linear least squares solver on the condensed points instead of the full data set, which reduces our cost per solve from $\mathcal{O}(\text{\lstinline{num_points}})$ to $\mathcal{O}(\text{\lstinline{num_samples}})$. This is purely a performance optimization, with no quality implications.
\begin{figure}[t]
  \centering
  \includegraphics[width=\linewidth]{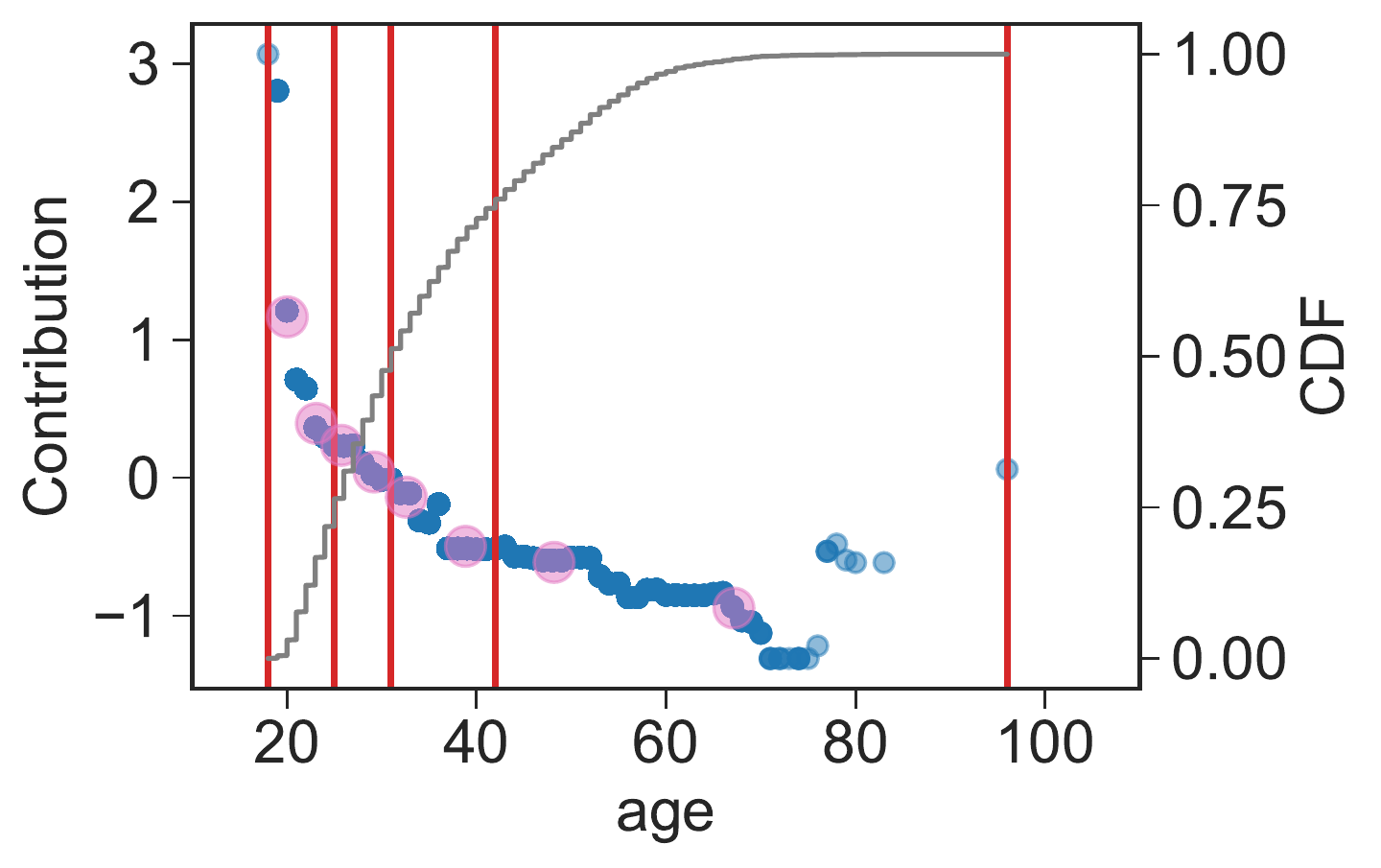}
  \caption{Candidate $x$-knots  (red vertical lines) and derived condensed points (pink large dots) on the \lstinline{age} piece of a COMPAS GAM forest model (blue dots). For visual clarity, this illustration considers only five $x$-knot candidates.}
  \label{fig:linear_condese}
\end{figure}
\subsubsection{Global Optimization via Greedy Search}
After discretization, the solution space consists of $\text{\lstinline{num_samples}} \choose {\text{\lstinline{num_segments}} + 1}$ $x$-knot combinations, which is still too large for an exhaustive search. To make the search tractable we use a greedy search heuristic that optimizes one $x$-knot at a time. Specifically, at each step of the process we evaluate the error associated with each candidate $x$-knot, and keep the candidate that yields the least error.

With this approach, we optimize in two stages. We begin with a single $x$-knot as our solution, and greedily add the best remaining candidate $x$-knot until our solution consists of \lstinline{(num_segments + 1)} $x$-knots. Then we cycle through our solution, removing one $x$-knot at a time and replacing that $x$-knot with the best remaining candidate $x$-knot, which could be the same $x$-knot that we just removed. We continue this cycle of iterative improvements until our solution converges, or until we've exceeded the maximum number of iterations (defaulting to 10 iterations).
\subsubsection{Slope Constraints \& Monotonicity}

\lstinline{pwlfit} can impose a minimum and/or maximum slope on the solution via bounded least squares. Instead of solving the least squares expression directly for the $y$-knots, we solve it for the deltas between adjacent $y$-knots. Then we impose a min/max slope by bounding the deltas. Slope restrictions can be used to limit the spikiness of curves, but we primarily use them to impose monotonicity. For example, specifying \lstinline{min_slope=0} restricts to monotonically non-decreasing functions while \lstinline{max_slope=0} restricts to monotonically non-increasing functions.  Specifying a \lstinline{min_slope} greater than 0 or a \lstinline{max_slope} less than 0 restricts to strictly increasing or decreasing functions, respectively.

\lstinline{pwlfit} can deduce the direction of monotonicity by applying isotonic regression \cite{wiki_Isotonic} to the condensed points. We fit an increasing and a decreasing isotonic regression, and use the direction that minimizes mean squared error. The user can override this behavior by specifying the direction explicitly or by disabling monotonicity entirely.

\subsubsection{Input Transformations}

\lstinline{pwlfit} can also interpolate in a transformed $x$-coordinate space instead of the original space, as a simple form of feature engineering. \lstinline{pwlfit} transforms the $x$-values before learning the curve.  Specifically, \lstinline{pwlfit} will choose a candidate $x$-transformation, \lstinline{fx}, among \lstinline{log}, \lstinline{log1p}, or \lstinline{symlog1p} based on the range of the $x$-values and then proceed with that transformation if it increases the Pearson correlation between \lstinline{fx} and $y$ by a noticeable amount over the identity transformation. Alternatively, the user can specify any strictly increasing 1D transform or specify the identity transform to disable transformation.

\section{Experiments}\label{sec:exp}

\subsection{Distillation Accuracy}
\begin{figure*}
\begin{subfigure}[t]{0.6\textwidth}
\centering
  \includegraphics[width=0.49\textwidth]{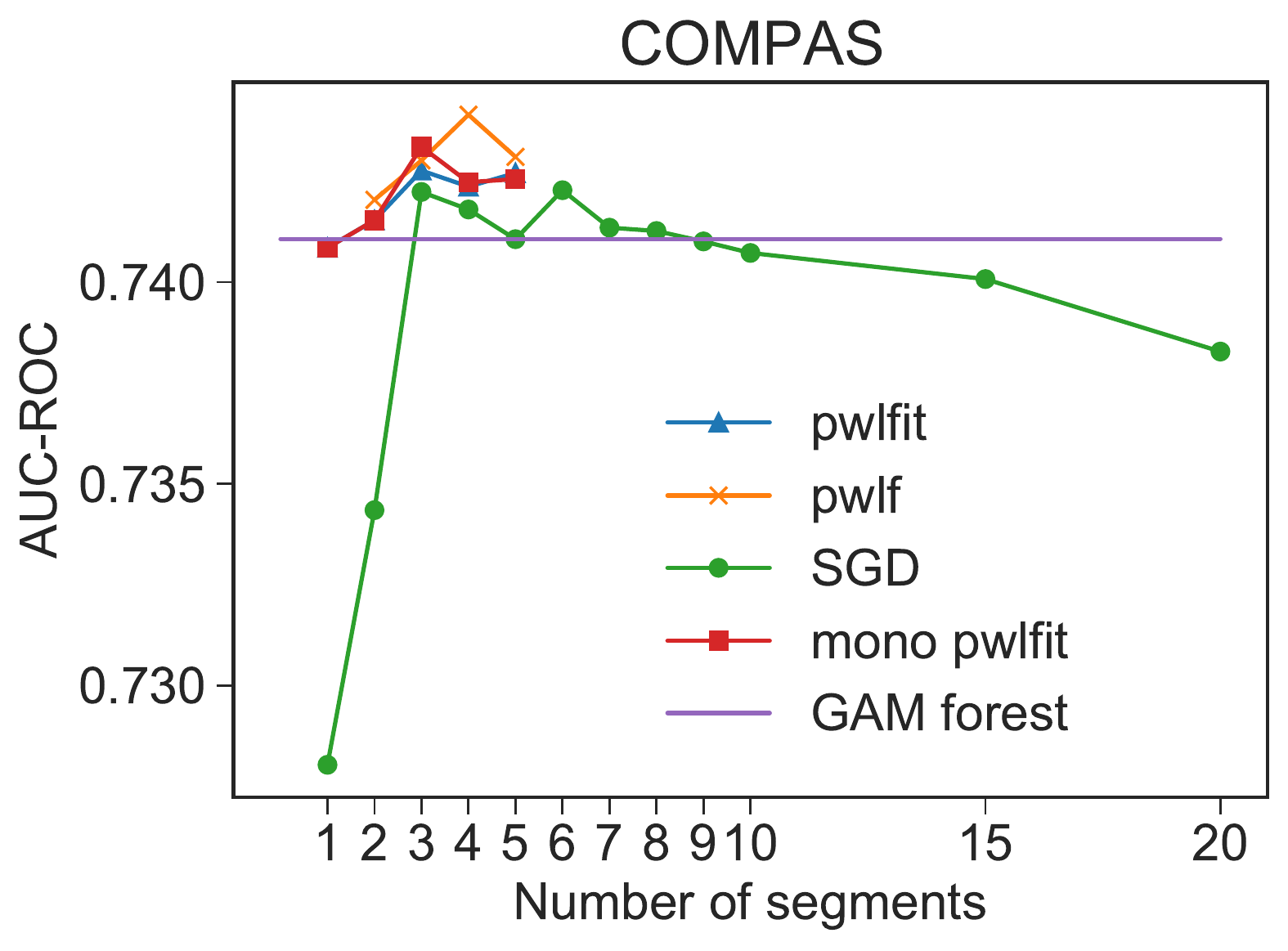}
  \includegraphics[width=0.49\textwidth]{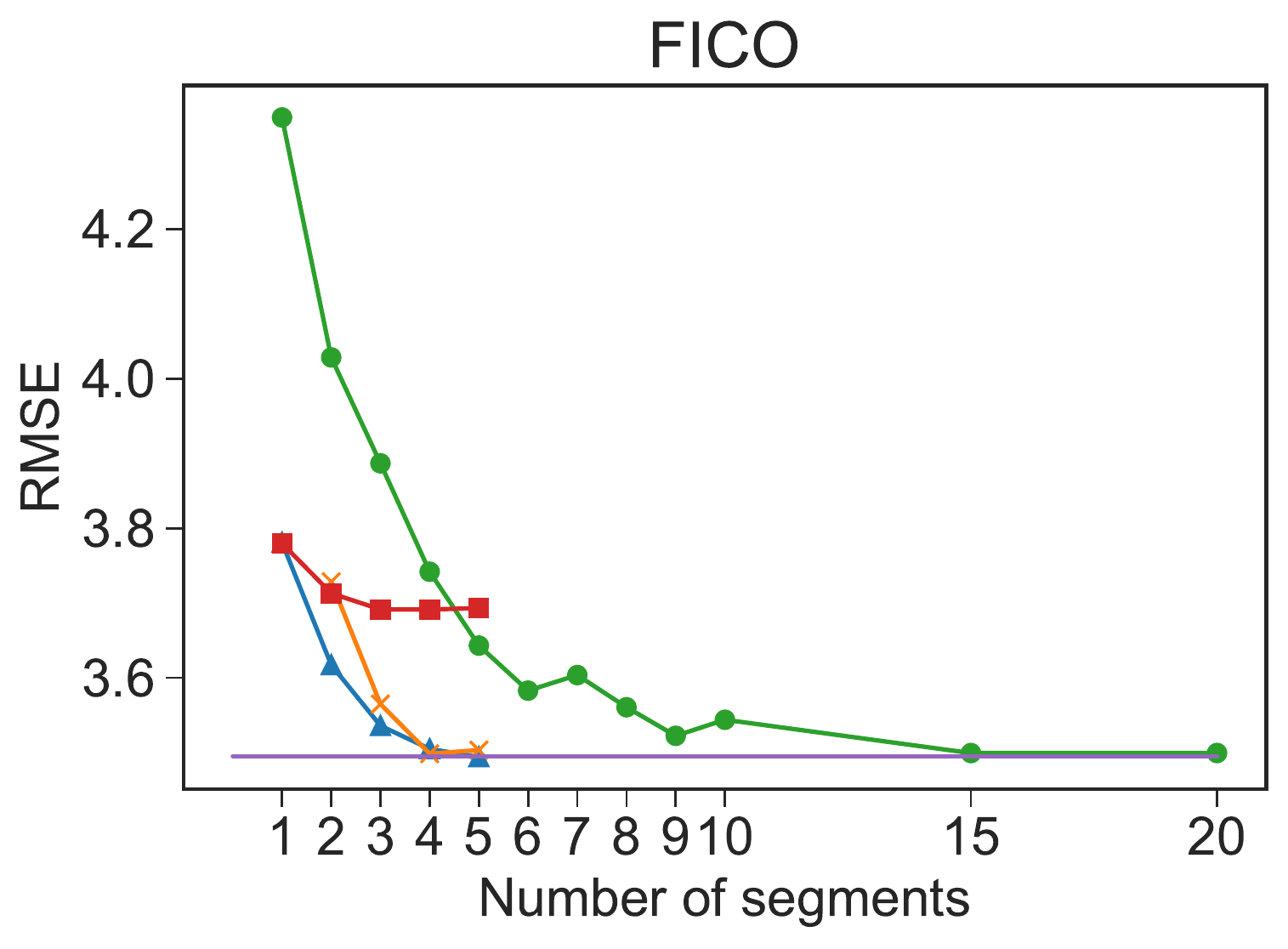}
  \includegraphics[width=0.49\textwidth]{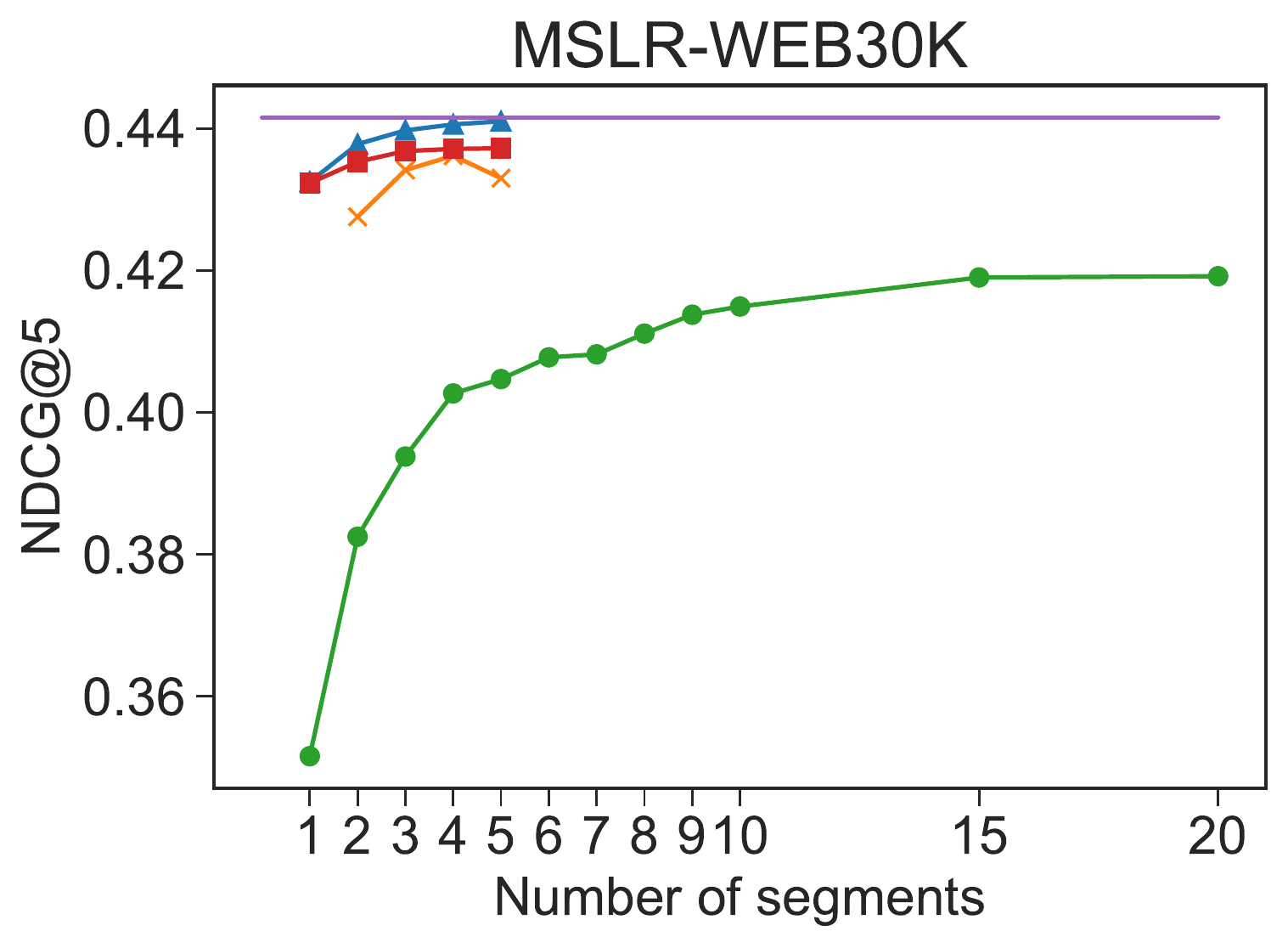}
  \includegraphics[width=0.49\textwidth]{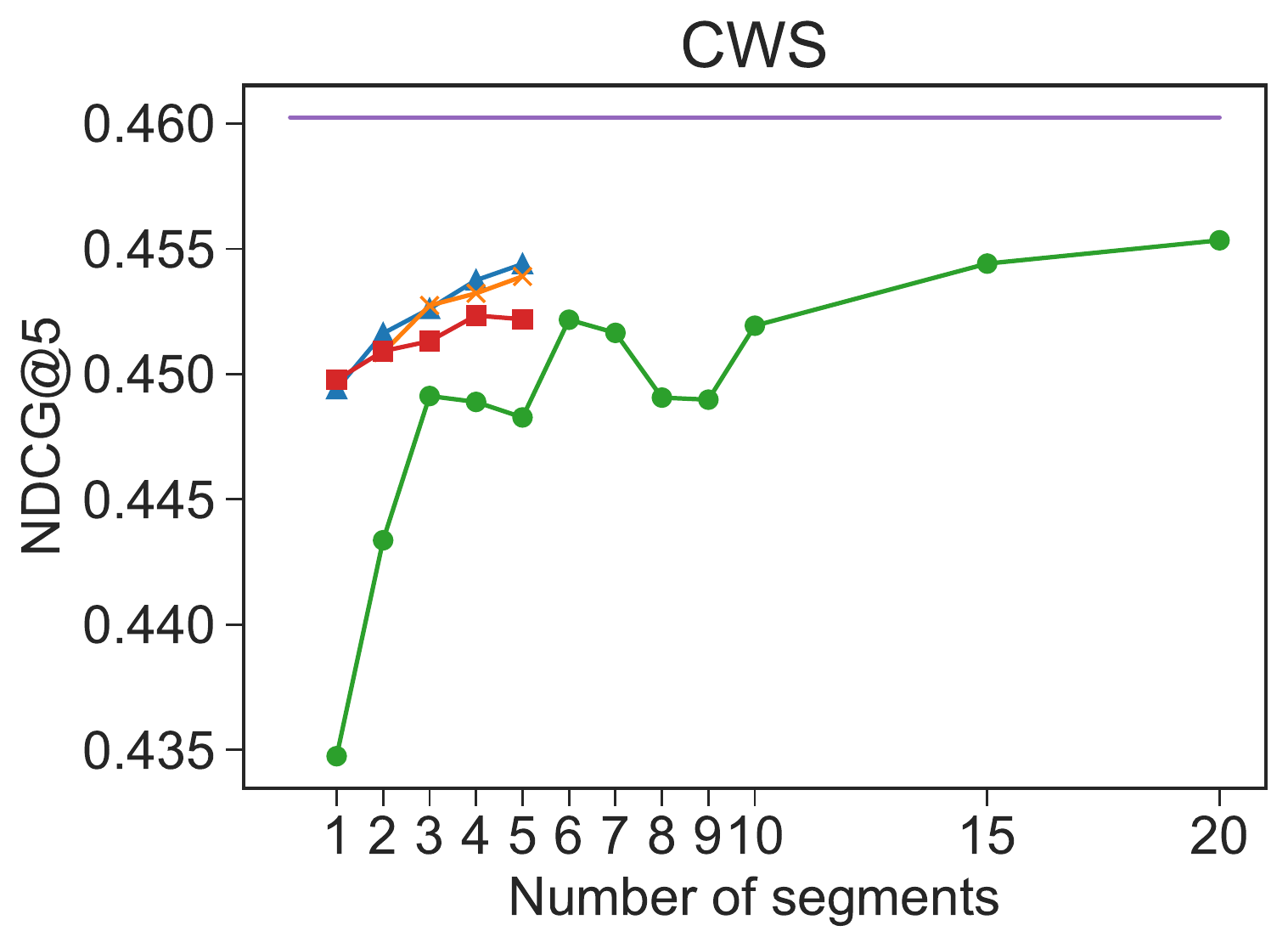}
  \caption{Top level metrics across datasets. For AUC-ROC and NDCG@5 higher is better, for RMSE lower is better.}
  \label{fig:aggregate_metrics}
\end{subfigure}
\hfill
\begin{subfigure}[t]{0.32\textwidth}
  \includegraphics[width=\textwidth]{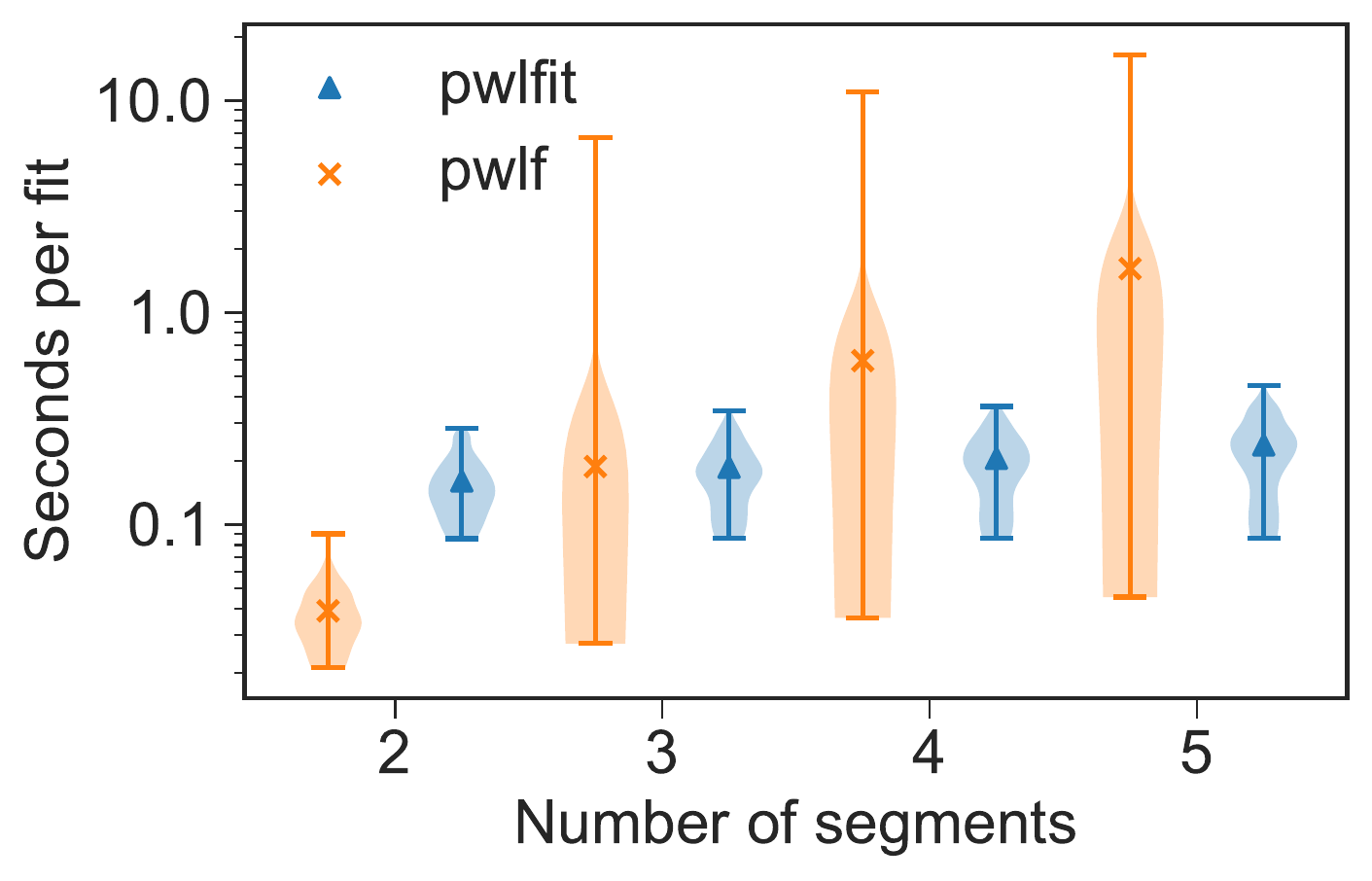}
  \includegraphics[width=\textwidth]{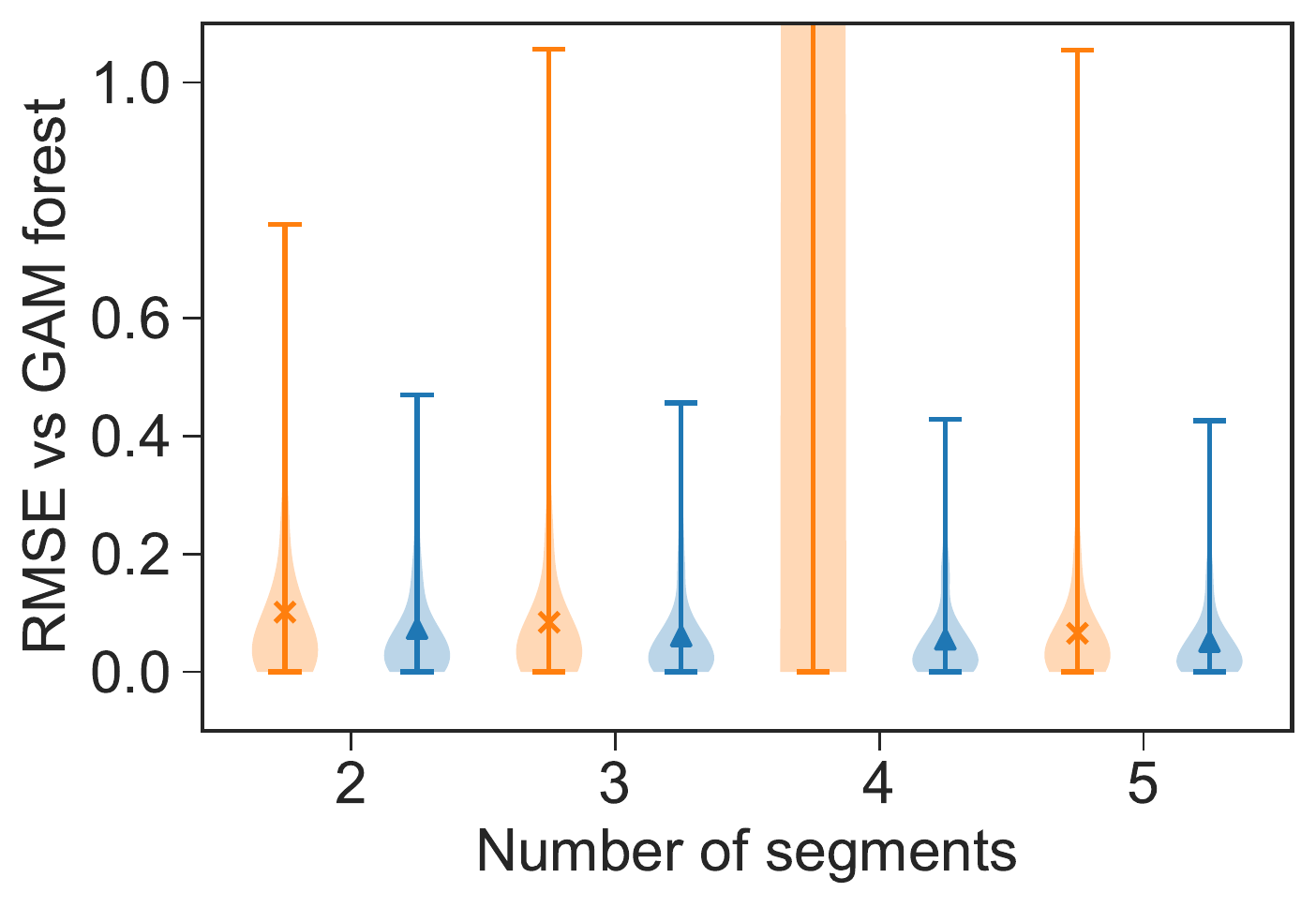}
  \caption{Per fit metrics for MSLR-WEB30K Fold 1. Note that for 4 segments \lstinline{pwlf} has extreme outliers for RMSE vs the source submodel.}
  \label{fig:per_fit_metrics}
\end{subfigure}
\caption{Comparisons of different methods across the 4 datasets.}
\end{figure*}

Table~\ref{tbl:short_aggregate_metrics} and Figure~\ref{fig:aggregate_metrics} show the results obtained from experiments on the different datasets. A complete set of results can be found in Table~\ref{tbl:long_aggregate_metrics} in Appendix~\ref{appendix:experiment_details}.  The results of applying our distillation technique with our piecewise-linear approximation algorithm are presented as \lstinline{pwlfit}. We present results from using various numbers of segments with and without a monotonicity restriction and otherwise default parameters. In all cases we truncated the control points to four significant digits. We also present several additional reference points to provide context.

\begin{itemize}
    \item \textbf{SGD:} We directly learn the curves with the Adadelta\cite{adadelta} optimizer. We initialize the $y$ values of the control points as zeros. For the $x$ values of the control points we use the quantiles for numerical features (\eg{ 0\%, 50\%, 100\% for a three point, two segment curve}) or all unique values for categorical features. We then apply Adadelta to optimize the $y$ values. Simultaneously optimizing $x$ and $y$ values was also attempted, but the results were always worse than optimizing $y$ values alone.
    \item \textbf{NAM:} Neural Additive Models (NAMs) \cite{nam} is another method for learning interpretable models proposed by Agarwal \etal. We present their result for reference where applicable. 
    \item \textbf{Interacting forest:} We train a bagged, boosted decision forest allowing feature interactions to demonstrate the accuracy of a non-interpretable, high-complexity "black box" model.
    \item \textbf{GAM forest}: We train a bagged boosted decision forest GAM by restricting each tree to use only one feature. This model is also the source model for our distillation technique.
    \item \textbf{\lstinline{pwlf}:} We apply our distillation technique using an alternative piecewise-linear approximation algorithm \lstinline{pwlf}\cite{pwlf}.
\end{itemize}

On each dataset we used five fold cross validation and present the metric mean and sample standard deviation across folds. We used three different metrics to evaluate accuracy: AUC-ROC, RMSE, and NDCG@5 for the three different tasks of classification, regression, and ranking. Further details on our experimentation setup can be found in Appendix~\ref{appendix:experiment_details} and further details on the datasets, labels, and metrics can be found in Preliminaries~\ref{sec:data-description}.

\begin{table*}
  \caption{Metrics across datasets. For AUC-ROC and NDCG@5 higher is better, and for RMSE lower is better. MSLR-WEB30K and CWS were not used by the NAM paper and are omitted from that row. Metric values are the mean from five fold cross validation $\pm$ the sample standard deviation.}
  \label{tbl:short_aggregate_metrics}
  \begin{tabular}{ccccc}
    \toprule
    Model & COMPAS (AUC-ROC) & FICO (RMSE) & MSLR-WEB30K (NDCG@5) &  CWS (NDCG@5)\\
    \midrule
Interacting forest & $0.742 \pm 0.011$ & $3.128 \pm 0.092$ & $0.485 \pm 0.002$ & $0.461 \pm 0.003$\\
GAM forest & $0.741 \pm 0.013$ & $3.495 \pm 0.109$ & $0.442 \pm 0.002$ & $0.460 \pm 0.004$\\
NAM & $0.741 \pm 0.009$ & $3.490 \pm 0.081$ & & \\
\lstinline[]$pwlfit num_segments=5, mono=False$ & $0.743 \pm 0.012$ & $3.494 \pm 0.096$ & $0.441 \pm 0.002$ & $0.454 \pm 0.002$\\
\lstinline[]$pwlfit num_segments=5, mono=True$ & $0.743 \pm 0.013$ & $3.693 \pm 0.101$ & $0.437 \pm 0.003$ & $0.452 \pm 0.003$\\
\lstinline[]$pwlf num_segments=5$ & $0.743 \pm 0.012$ & $3.503 \pm 0.096$ & $0.433 \pm 0.003$ & $0.454 \pm 0.003$\\
SGD num\_segments=5 & $0.741 \pm 0.010$ & $3.643 \pm 0.097$ & $0.405 \pm 0.002$ & $0.448 \pm 0.004$\\
SGD num\_segments=20 & $0.738 \pm 0.011$ & $3.499 \pm 0.117$ & $0.419 \pm 0.003$ & $0.455 \pm 0.003$\\
  \bottomrule
\end{tabular}
\end{table*}

Our results show that applying our distillation technique with 4-5 segments with \lstinline{pwlfit} produces models which are as accurate as both the source GAM forest and NAM models for all datasets except CWS where a small gap remains. We investigate this accuracy gap in detail in Section~\ref{subsec:distillation-failures} below. In the case of the COMPAS dataset these models are as accurate as full complexity models. Applying our technique with \lstinline{pwlf} produces competitive results, albeit less accurate on the MSLR-WEB30K dataset. By contrast, the results show that learning curves directly via SGD is less general. On the FICO and CWS datasets more segments are required to achieve accuracy comparable to the GAM forest models. On the MSLR-WEB30K dataset the accuracy is inferior even with many more segments.

The consistent accuracy of applying our distillation approach with \lstinline{pwlfit} on these four datasets and three separate tasks (classification, regression, learning to rank) demonstrates that the process is not sensitive to either the specific data or the top level objective being used.

\subsection{Distillation Failures}
\label{subsec:distillation-failures}
\begin{figure}[t]
  \includegraphics[width=\linewidth]{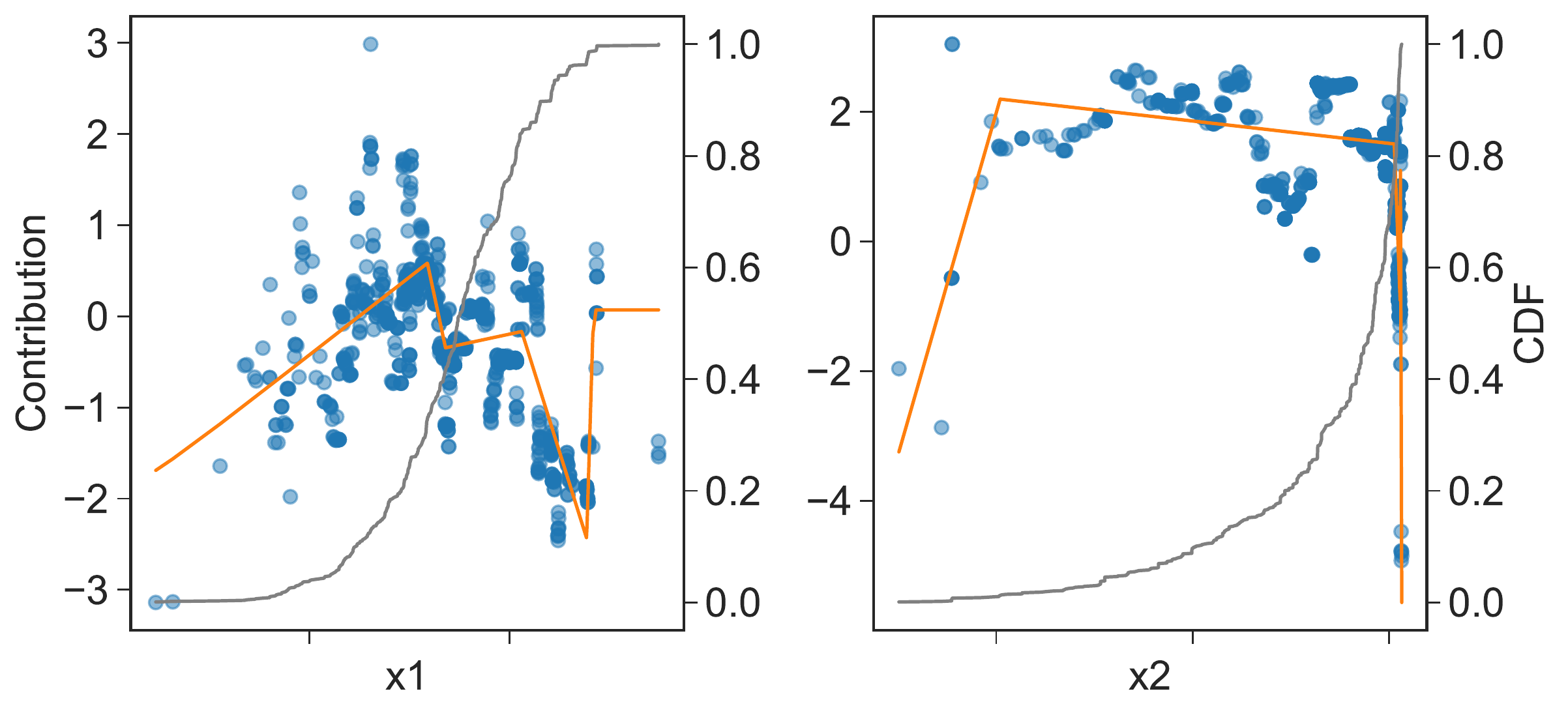}
\caption{The worst curve distillations from the CWS dataset using 5 segments.}
\label{fig:cws_bad_fit}
\end{figure}

In Section~\ref{subsec:greater-transparency} we explained how distillation yielding a model with inferior accuracy warrants further investigation because the purported "failure" can often be attributed to essential yet non-interpretable characteristics of the teacher model not transferring to the student model. The accuracy gap observed on the CWS dataset is an example of this phenomenon. Figure~\ref{fig:cws_bad_fit} shows the worst two fits from the CWS dataset. The plots have been redacted to maintain the privacy of the dataset. For each plot it is clear that the original teacher submodel had some non-interpretable behavior which was lost during distillation. This is most evident for feature $x_1$, the worst offender, where the output is highly erratic. If the original teacher submodel is not distilled for these two features then the accuracy gap between the original teacher model and 5 segment non-monotonic distillation drops from 0.0059 to 0.003 (\ie{ \textasciitilde{50\%} of the gap is recovered}). 

To identify the above two failures we applied the following method.
\begin{itemize}
    \item Begin with the original teacher model. For each submodel compute the metric delta against the teacher model from distilling only that submodel and no others.
    \item Perform the above on each cross validation fold using the validation set and average the metric deltas across folds.
    \item Sort the features by their associated metric delta to determine the worst distillations.
\end{itemize}

\subsection{Efficiency \& Robustness}
The experiments of the previous section showed that \lstinline{pwlfit} more accurately distills the source model across datsets than \lstinline{pwlf}. We also found on the MSLR-WEB30K dataset that \lstinline{pwlfit} is more efficient and robust than \lstinline{pwlf}.  Figure~\ref{fig:per_fit_metrics} shows per fit metrics from the first fold of the MSLR-WEB30K dataset as the number of segments varies without monotonicity. The top plot shows the time in seconds, as measured on a ThinkStation P520, to fit each of the 136 submodels of the source GAM forest. We find that \lstinline{pwlfit} is faster in the average case as the number of segments increases, and has a narrower distribution. The bottom plot shows the RMSE of each fit against the 136 submodels of the source GAM forest. We again find that \lstinline{pwlfit} performs favorably in the average case with a narrower runtime distribution.

It's worth noting that \lstinline{pwlf} by default does not perform any downsampling. For the MSLR-WEB30K dataset running \lstinline{pwlf} without any downsampling was prohibitively expensive. For all of our experiments we ran \lstinline{pwlf} with a pre-processing downsample to a random 1000 examples. We found this to be a fair point for balancing speed and quality when comparing to \lstinline{pwlfit}. It is of course possible with both algorithms to modify the number of samples used to strike a different trade-off between run time and accuracy.

\subsection{Monotonicity}
\label{subsec:Monotonicity}

As discussed in Section~\ref{sec:Piecewise-Linear Curve Approximation}, \lstinline{pwlfit} can fit monotonic curves with automatic direction detection. Figure~\ref{fig:aggregate_metrics} compares curve models fit with and without monotonicity constraints (automatically inferring the direction) across datasets. For the COMPAS dataset monotonic and non monotonic models are comparably accurate, while for FICO, MSLR-WEB30K, and CWS, non-monotonic models are more accurate.

\begin{figure}[t]
  \includegraphics[width=\linewidth]{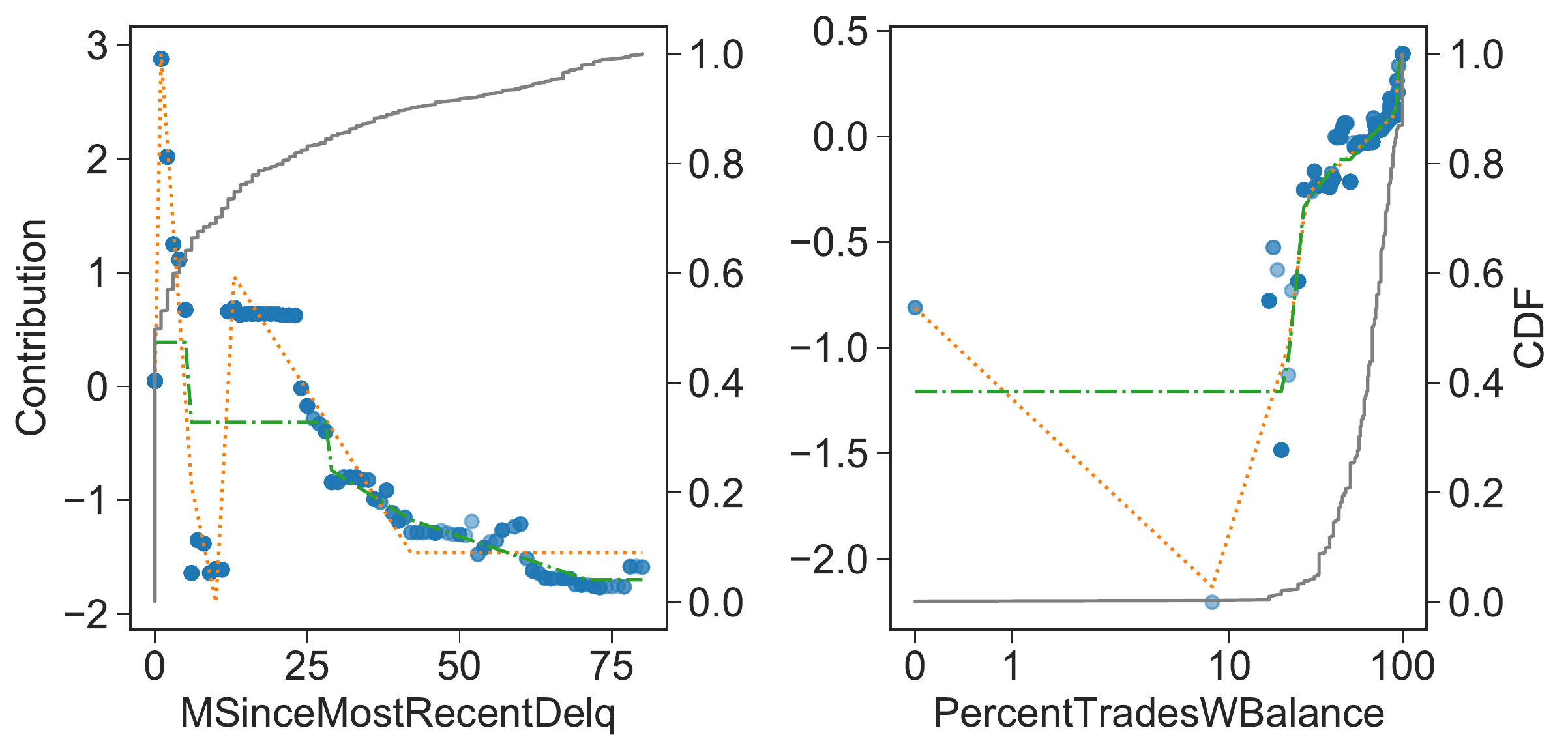}
\caption{Curve distillations on the FICO dataset using 5 segments without monotonicity (orange) and with monotonicity (green).}
\label{fig:FICO_mono_vs_non_mono}
\end{figure}

Monotonicity with respect to appropriate features is desirable for interpretable models. In these cases a monotonic model may be preferable to a non-monotonic one, even if it is less accurate. For example, Figure~\ref{fig:FICO_mono_vs_non_mono} compares monotonic and non-monotonic 5 segment curve models on the FICO dataset for the \lstinline{MSinceMostRecentDelq}, and \lstinline{PercentTradesWBalance} features. Given the semantic meaning of these features, it is desirable from a transparency and incentives standpoint for the model output to be monotonic with respect to each of them.

\section{Conclusion}\label{sec:conclusion}
We have introduced a novel method for distilling interpretable models into human-readable code using piecewise-linear curves and demonstrated its efficacy on four datasets.  We have shown that curve models match or outperform the accuracy achieved by other additive models. On smaller datasets, curve models match the accuracy of more complex models, like interacting decision forests.  Our localized distillation methodology is applicable to any model containing univariate numerical functions and is straightforward to implement using the publicly-available \lstinline{pwlfit}\cite{pwlfit} library.

We have explained how curve model distillation reinforces interpretability and addresses a variety of real-world engineering challenges. Curve models are 1) transparent, 2) well-regularized, 3) easy to analyze for presence of biases or other fairness issues, and 4) can be directly edited or improved outside the context of machine learning to fix the aforementioned fairness issues. Distilling models into human-readable code allows one to address novel machine learning problems using well-established software engineering methods. Curve models can be improved by multiple contributors in parallel, reviewed, and made to systematically follow best practices. Curve models are well-suited for production applications, since they can be natively supported in many languages, are easy to deploy, and fast to evaluate.

\begin{acks}
We thank Vytenis Sakenas, Jaime Fernandez del Rio, Benoit Zhong, and Petr Mitrichev for their support and providing the algorithms and optimization infrastructure used in our experiments. We also thank Paul Heymann, Diego Federici, Mike Bendersky, Paul Haahr, and Petr Mitrichev for their helpful feedback and detailed reviews. Lastly, we thank Xinyu Qian, Janelle Lee, Po Hu, and Chary Chen for preparing the CWS data set for our experiments.
\end{acks}

\bibliographystyle{ACM-Reference-Format}
\bibliography{main}

\clearpage % Note KDD requests that extra material start on page 10. Hence the clear page. 
\appendix
\section{Experimental Details}
\label{appendix:experiment_details}
\begin{table*}
  \caption{Metrics across datasets. For AUC-ROC and NDCG@5 higher is better, and for RMSE lower is better. MSLR-WEB30K and CWS were not used by the NAM paper and are omitted from that row. Metric values are the mean from five fold cross validation $\pm$ the sample standard deviation.}
  \label{tbl:long_aggregate_metrics}
  \begin{tabular}{ccccc}
    \toprule
    Model & COMPAS (AUC-ROC) & FICO (RMSE) & MSLR-WEB30K (NDCG@5) &  CWS (NDCG@5)\\
    \midrule
Interacting forest & $0.742 \pm 0.011$ & $3.128 \pm 0.092$ & $0.485 \pm 0.002$ & $0.461 \pm 0.003$\\
GAM forest & $0.741 \pm 0.013$ & $3.495 \pm 0.109$ & $0.442 \pm 0.002$ & $0.460 \pm 0.004$\\
NAM & $0.741 \pm 0.009$ & $3.490 \pm 0.081$ & & \\
\hline
\lstinline[]$pwlfit num_segments=1, mono=False$ & $0.741 \pm 0.008$ & $3.781 \pm 0.105$ & $0.432 \pm 0.003$ & $0.449 \pm 0.003$\\
\lstinline[]$pwlfit num_segments=1, mono=True$ & $0.741 \pm 0.008$ & $3.780 \pm 0.105$ & $0.432 \pm 0.003$ & $0.450 \pm 0.003$\\
\lstinline[]$pwlfit num_segments=2, mono=False$ & $0.742 \pm 0.011$ & $3.617 \pm 0.101$ & $0.438 \pm 0.002$ & $0.452 \pm 0.003$\\
\lstinline[]$pwlfit num_segments=2, mono=True$ & $0.742 \pm 0.011$ & $3.713 \pm 0.103$ & $0.435 \pm 0.003$ & $0.451 \pm 0.003$\\
\lstinline[]$pwlfit num_segments=3, mono=False$ & $0.743 \pm 0.010$ & $3.536 \pm 0.099$ & $0.440 \pm 0.002$ & $0.453 \pm 0.002$\\
\lstinline[]$pwlfit num_segments=3, mono=True$ & $0.743 \pm 0.011$ & $3.691 \pm 0.100$ & $0.437 \pm 0.002$ & $0.451 \pm 0.003$\\
\lstinline[]$pwlfit num_segments=4, mono=False$ & $0.742 \pm 0.012$ & $3.505 \pm 0.094$ & $0.441 \pm 0.002$ & $0.454 \pm 0.003$\\
\lstinline[]$pwlfit num_segments=4, mono=True$ & $0.742 \pm 0.012$ & $3.691 \pm 0.101$ & $0.437 \pm 0.003$ & $0.452 \pm 0.004$\\
\lstinline[]$pwlfit num_segments=5, mono=False$ & $0.743 \pm 0.012$ & $3.494 \pm 0.096$ & $0.441 \pm 0.002$ & $0.454 \pm 0.002$\\
\lstinline[]$pwlfit num_segments=5, mono=True$ & $0.743 \pm 0.013$ & $3.693 \pm 0.101$ & $0.437 \pm 0.003$ & $0.452 \pm 0.003$\\
\hline
\lstinline[]$pwlf num_segments=2$ & $0.742 \pm 0.014$ & $3.728 \pm 0.101$ & $0.428 \pm 0.004$ & $0.451 \pm 0.003$\\
\lstinline[]$pwlf num_segments=3$ & $0.743 \pm 0.012$ & $3.565 \pm 0.108$ & $0.434 \pm 0.004$ & $0.453 \pm 0.003$\\
\lstinline[]$pwlf num_segments=4$ & $0.744 \pm 0.012$ & $3.498 \pm 0.099$ & $0.436 \pm 0.004$ & $0.453 \pm 0.003$\\
\lstinline[]$pwlf num_segments=5$ & $0.743 \pm 0.012$ & $3.503 \pm 0.096$ & $0.433 \pm 0.003$ & $0.454 \pm 0.003$\\
\hline
SGD num\_segments=1 & $0.728 \pm 0.008$ & $4.349 \pm 0.059$ & $0.352 \pm 0.002$ & $0.435 \pm 0.003$\\
SGD num\_segments=2 & $0.734 \pm 0.007$ & $4.028 \pm 0.095$ & $0.382 \pm 0.001$ & $0.443 \pm 0.004$\\
SGD num\_segments=3 & $0.742 \pm 0.008$ & $3.887 \pm 0.080$ & $0.394 \pm 0.002$ & $0.449 \pm 0.003$\\
SGD num\_segments=4 & $0.742 \pm 0.010$ & $3.742 \pm 0.110$ & $0.403 \pm 0.003$ & $0.449 \pm 0.004$\\
SGD num\_segments=5 & $0.741 \pm 0.010$ & $3.643 \pm 0.097$ & $0.405 \pm 0.002$ & $0.448 \pm 0.004$\\
SGD num\_segments=6 & $0.742 \pm 0.010$ & $3.583 \pm 0.105$ & $0.408 \pm 0.002$ & $0.452 \pm 0.002$\\
SGD num\_segments=7 & $0.741 \pm 0.010$ & $3.604 \pm 0.098$ & $0.408 \pm 0.002$ & $0.452 \pm 0.003$\\
SGD num\_segments=8 & $0.741 \pm 0.010$ & $3.561 \pm 0.111$ & $0.411 \pm 0.003$ & $0.449 \pm 0.004$\\
SGD num\_segments=9 & $0.741 \pm 0.011$ & $3.522 \pm 0.103$ & $0.414 \pm 0.002$ & $0.449 \pm 0.003$\\
SGD num\_segments=10 & $0.741 \pm 0.010$ & $3.544 \pm 0.117$ & $0.415 \pm 0.002$ & $0.452 \pm 0.005$\\
SGD num\_segments=15 & $0.740 \pm 0.011$ & $3.499 \pm 0.101$ & $0.419 \pm 0.003$ & $0.454 \pm 0.003$\\
SGD num\_segments=20 & $0.738 \pm 0.011$ & $3.499 \pm 0.117$ & $0.419 \pm 0.003$ & $0.455 \pm 0.003$\\
  \bottomrule
\end{tabular}
\end{table*}

\subsection{Cross Validation} We performed 5-fold cross validation on all datasets. 
\begin{itemize}
    \item \textbf{COMPAS \& FICO}: The datasets were split into 5 equal parts. Each part was used once as a test set (20\%) with the remaining parts as the training set (80\%). We used the same random folds as in the NAM paper \cite{nam}. No validation set was used given the small size of the data. Instead we used out of bag evaluation wherever a validation set would be used (see below). 
    \item \textbf{MSLR-WEB30K}: We used the predefined folds and partitions from the original dataset. For each fold it allocates 60\% for training 20\% for validation and 20\% for testing.
    \item \textbf{CWS}: We used a dataset of 60,000 queries and 2,690,439 items with an average of \textasciitilde{44} items per query. The dataset was split into 5 equal parts. Each part was used once as a test set. Of the remaining parts 80\% was used as training and 20\% as validation. Overall this resulted in 64\% for training, 16\% for validation and 20\% for test for each fold. 
\end{itemize}

\subsection{Ensemble Learning}
For both SGD and tree models, we trained ensembles with 56 bags using a bag fraction of $\frac{7}{8}$ to produce the random subsets. For MSLR-WEB30K and CWS, queries were randomly divided into bags. For the other datasets, individual examples were randomly divided into bags. When applying our distillation technique, we distilled the ensembles into a single \lstinline{PWLCurve} per feature. When learning the curves directly via SGD, we averaged the learned $y$-coordinate values across bags to obtain the final model.

\subsection{Loss Functions}
We trained SGD and tree models using log-loss for the COMPAS dataset, mean squared error (MSE) for the FICO dataset, and ranking loss (ApproxNDCG~\cite{qin2010general}) for the MSLR-WEB30K and CWS datasets.

\subsection{Hyper-parameters}
For the COMPAS, and FICO datasets hyper-parameters were tuned using out of bag evaluation on the training set of the first fold. For MSLR-WEB30K and CWS, we used the validation sets of the first fold.
\begin{itemize}
    \item \textbf{SGD:} We tuned the batch size in \{128, 256, 512, 1024, 4096\}. We used the Adadelta \cite{adadelta} optimizer and tuned a sufficient maximum number of steps for convergence. No other parameters were tuned.
    \item \textbf{Interacting forest:} We trained depth 5 trees using an internal boosted forest algorithm. We tuned a sufficient maximum number of steps for convergence. No other parameters were tuned.
    \item \textbf{GAM forest:} We trained depth 3 trees restricted to using a single feature with an internal boosted forest algorithm. We tuned a sufficient maximum number of steps for convergence. No other parameters were tuned.
\end{itemize}

In all cases, we trained models for the tuned maximum number of steps and then truncated models after training. Truncation used a confidence-based truncation algorithm which attempts to select the earliest step for which no later step provides a confident win. This algorithm was run on the validation set if present or otherwise utilized out of bag evaluation.

\subsection{Code}
The GitHub repository for \lstinline{pwlfit} \cite{pwlfit} contains several Jupyter notebooks applying our distillation technique and performing the analyses shown in this paper. Please reference the v0.2.0 release to get the accompanying data files and appropriate version of the Jupyter notebooks.

\FloatBarrier
\section{Linear Condense}
\label{appendix:linear_condense}

Linear condensing is a data optimization designed to reduce the runtime complexity of our piecewise-linear curve fitting.

\subsection{Motivation/Overview}

\lstinline{pwlfit} picks a set of candidate $x$-knots and searches through combinations of those $x$-knots. For each combination considered, it solves a linear least squares expression for the ideal $y$-knots, calculates the resulting squared error, and prefers the combination that yields the lowest error.

Each solve is linear in the size of input, which is slow for large data. We could downsample to save compute at the cost of accuracy. Instead, we introduce a technique to save compute at no cost in accuracy. We condense the data into $\mathcal{O}(\#candidates)$ synthetic points. These synthetic points perfectly recreate the true squared error over the full data for every \lstinline{PWLCurve} that will be considered. We then optimize over the synthetic points instead of the real points.

This is possible because we know the candidate $x$-knots ahead of time. A \lstinline{PWLCurve} defined on those $x$-knots will always be linear between any adjacent $x$-knots in the set of candidates. As we show in the theorem, we can condense arbitrarily many points down to two points such that linear fits are the same on those two points as on the full set. In the corollary, we apply this process separately between each pair of candidate $x$-knots, producing two points between each pair. Together, the squared error of such a \lstinline{PWLCurve} is the same on those synthetic points as it is on the full data set. (Up to a constant that we safely ignore because it's the same for each \lstinline{PWLCurve}.)

\subsection{Definitions}
For convenience, we take standard definitions and specialize them for weighted 2D points.

\begin{definition}
Let a `point' refer to a real-valued triple of the form $(x, y, weight)$ with $weight > 0$.
\end{definition}

\begin{definition}
Let a `line' refer to a function of the form $f(x) = mx+b$ for $m,b,x \in \mathbb{R}$.
\end{definition}

\begin{definition}
For any function $f:\mathbb{R} \to \mathbb{R}$ and finite point set $P$, define the squared error $SE(f, P)$ as the sum of $(f(x) - y)^2 \cdot weight$ for each point in $P$. If $P$ is empty, we consider the squared error to be $0$.
\end{definition}

\begin{definition}
For any finite point set $P$, define the `best fit line' $best fit line(P)$ as the line $L$ that minimizes $SE(L, P)$. In the degenerate case where multiple lines minimize $SE$, let the best fit line be the solution with zero slope, and if multiple solutions have zero slope, let the best fit line be the solution with a zero $y$-intercept.
\end{definition}

There are two degenerate cases that require tie-breaking. If the point set is empty, every line has the same squared error, so our definition chooses $f(x)=0$ as the best fit line. If the point set is nonempty but all its points have the same $x$, then any line with the correct value at $x$ will minimize the squared error, so our definition choose the horizontal line.

\subsection{Theorem}
\begin{theorem}
Given a set of points $P$, we can construct a set $P'$ of two or fewer points such that

1.   $min_x(P) <= min_x(P') <= max_x(P') <= max_x(P)$, and

2.   For any line $L$, $SE(L, P) = SE(L, P') + SE(best fit line(P), P)$.
\end{theorem}

\theoremstyle{remark}
\newtheorem*{remark}{Remark}

\begin{remark}
These properties are desirable because (2) allows us to compute the squared error of $M$ lines over a data set of $N$ points in $\mathcal{O}(N+M)$ instead of the naive $\mathcal{O}(NM)$, and (1) allows us to extend this property from lines to a useful class of piecewise-linear curves in the corollary.

Note that the points in $P'$ are constructed, rather than chosen from $P$. The construction of $P'$ is implemented in \lstinline{pwlfit} \cite{pwlfit} as \lstinline{linear_condense.linear_condense}.
\end{remark}

\begin{proof}
Let $X$, $Y$, and $W$ represent the $x, y$, and $weight$ values of $P$, respectively. We dismiss the trivial case where $P$ is empty; in that case, an empty $P'$ satisfies the requirements. Likewise, we dismiss the case where $min(X) = max(X)$ since $P'=\{Centroid(P)\}$ fulfills our desired properties. With those cases resolved, we assume for the rest of this proof that $min(X) < max(X)$.

\subsubsection{Reframe the Coordinate System}
To begin, we reframe the coordinate system such that the origin is the centroid of $P$ and $y=0$ is the best fit line. (This simplifies the math.) We ensure that the shift of coordinates is reversible and preserves the squared error.

$Centroid(P) = (X \cdot W / sum(W), Y \cdot W / sum(W))$.
We translate the coordinate frame by this centroid so that, under the new coordinates, $Centroid(P) = (0,0)$. After translation, $X \cdot W = 0$ and $Y \cdot W = 0$.

Additionally, we skew the coordinate system by the slope of the best fit line: we replace $Y$ with $Y - X \cdot slope(best fit line(P))$. With the centroid at the origin, the slope of the best fit line is $Covariance(X, Y, W) / Variance(X, W)$ = $(XY \cdot W) / (XX \cdot W)$. After skewing this slope to 0, $XY \cdot W$ = 0.

Under the new coordinate frame, $SE(best fit line(P), P) = SE(y=0, P) = Y^2 \cdot W$.

We will determine $P'$ under this new coordinate system. Afterwards, we can easily convert $P'$ back to the original coordinate system by reversing the skew and the translation.

\subsubsection{Squared Error of an arbitrary line}
We will express $SE(line, P)$ as $SE(best fit line(P), P)$ plus leftover terms. From that, we will derive a $P'$ such that $SE(line, P')$ equals those leftover terms.

For an arbitrary line $y=mx+b$,

$SE(y=mx+b, P) = (mX + b - Y)^2 \cdot W = (m^2 X^2 + 2bmX - 2 mXY + b^2 - 2bY + Y^2) \cdot W.$

In our coordinate frame, $X \cdot W = 0$, $Y \cdot W = 0$, and $XY \cdot W = 0$. So $SE(y=mx+b, P) = (m^2 X^2 + b^2 + Y^2) \cdot W.$

$Y^2 \cdot W = SE(best fit line(P), P)$. Therefore, $$SE(y=mx+b, P) = m^2 X^2 \cdot W + b^2 \cdot W + SE(best fit line(P), P).$$
$$m^2 X^2 \cdot W + b^2 \cdot W = SE(y=mx+b, P) - SE(best fit line(P), P).$$

\subsubsection{Squared error over $P'$}

$$SE(y=mx+b, P') = SE(y=mx+b, P) - SE(best fit line(P), P)$$ for all lines $y=mx+b$ $\iff$ $(mX' + b - Y')^2 \cdot W' = m^2 X^2 \cdot W + b^2 \cdot W$ for all lines $y=mx+b$.

The above equation can be viewed as a quadratic polynomial in the two variables $m$ and $b$. To hold for all values of $m$ and $b$, the coefficients of each $m^c b^d$ must be equal on both sides of the equation. Then the equation holds iff:

1.   $X'^2 \cdot W' = X^2 \cdot W$, and

2.   $X' \cdot W' = 0$, and

3.   $sum(W) = sum(W')$, and

4.   $Y' \cdot W' = 0$, and

5.   $Y'^2 \cdot W' = 0$, and

6.   $X'Y' \cdot W' = 0$.

(5) $\iff$ $Y' = 0$, which also guarantees (4) and (6). We will use 1-3 to derive a satisfactory $X'$ and $W'$.

\subsubsection{Deriving $X'$ and $W'$}
We've determined that $Y' = 0$.

Let $X' := (x_1, x_2)$ and $W' := (w_1, w_2)$. Without loss of generality, let $x_1$ <= $x_2$. Then, to satisfy our squared error expression, it's necessary and sufficient that:

1.   $x_1^2 w_1 + x_2^2 w_2 = X^2 \cdot W$, and

2.   $x_1 w_1 + x_2 w_2 = 0$, and

3.   $w_1 + w_2 = sum(W)$.

Because we have three equations in four unknowns, we cannot directly solve for $x_1, x_2, w_1, w_2.$ To produce a fourth equation, we choose the constraint that $x_1 / x_2$ = $min(X) / max(X)$. This choice will simplify the math, and will ensure that $min(X) <= x_1 <= x_2 <= max(X)$.

With this fourth equation, we solve the simultaneous equations to produce:

$x_1 = -stddev(X, W) \sqrt{-min(X) / max(X)}$

$x_2 = stddev(X, W) \sqrt{max(X) / -min(X)}$.

$w_1 = sum(W) \cdot max(X) / (max(X) - min(X))$

$w_2 = sum(W) \cdot -min(X) / (max(X) - min(X))$.

Note that, because the centroid is zero, $min(X) < 0 < max(X)$, so these expressions are all defined. (The denominators are never 0 and values beneath the square roots are never negative.)

$P' = {(x_1, 0, w_1), (x_2, 0, w_2)}$ satisfies our requirements.

\subsubsection{Verify that $min(X) <= x_1 <= x_2 <= max(X)$}

We wanted $P'$ to satisfy the the squared error expression, which it does, and also have its x-values bounded by the x-values of $P$, which we prove now. Let $\mu:= E(X, W)$, the expected value of $X$ weighted by $W$, which is equivalent to the x-value of $Centroid(P)$. By the Bhatia–Davis inequality \cite{wiki_Bhatia_Davis},

$stddev(X,W)^2 <= (\mu - min(X))(max(X) - \mu)$. (This inequality is equivalent to the observation that the standard deviation of a distribution is maximized when all the xs are at the extremes -- i.e. equal to min(X) or max(X).)

Since $\mu$ is zero for $P$, $stddev(X,W)^2 <= -min(X)max(X)$.

$x_1^2 = stddev(X,W)^2 \cdot (-min(X) / max(X)) <= -min(X)max(X) \cdot (-min(x) / max(X)) = min(X)^2.$

$x_1 < 0$ and $min(X) < 0$, so $x_1^2 <= min(X)^2 \implies min(X) <= x_1$. The proof that $x_2 <= max(X)$ is similar.

Therefore $min(X) <= x_1 <= x_2 <= max(X)$, as desired.
\end{proof}

\subsection{Corollary}
\begin{corollary}
Given a set of points $P$ and a set of x-knots $K$, we can construct a set of points $P'$ with $|P'| <= 2(|K|- 1)$ such that, for any \lstinline{PWLCurve} $C$ whose x-knots are elements of $K$, $SE(C, P) = SE(C, P') + c$, where $c$ is a constant determined exclusively by $P$ and $K$ that's the same for every $C$.
\end{corollary}

Note that the points in $P'$ are constructed, rather than chosen from $P$. The construction of $P'$ is implemented in \lstinline{pwlfit} \cite{pwlfit} as \lstinline{linear_condense.condense_around_knots}. 
% \url{https://github.com/google/pwlfit} as linear\_condense.condense\_around\_knots.

\subsubsection{Preprocess $P$ by clamping}

Let $k:= |K|$, and consider $K$ in sorted order. Piecewise-linear curves are constant for input values that exceed the range of their x-knots, so $C$ is constant for $x <= min(K) = K[0]$ and for $x >= max(K) = K[k-1]$. 

Therefore we can clamp the x-values of $P$ to $[K[0], K[k-1]]$ without altering $SE(C, P)$. We do so as a preprocess.

\subsubsection{Partition $P$ by $K$}

To construct $P'$ from $P$, we first partition $P$ by $K$ into $k+1$ disjoint pieces labeled $P_0$, $P_1$, ..., $P_k$.

- $P_0$ := $\{x \in P | x < K[0]\}$.

- $P_i$ := $\{x \in P | K[i-1] <= x < K[i]\}$ for $1 <= i <= k-2$.

- $P_{k-1}$ := $\{x \in P | K[k-2] <= x <= K[k-1]\}$.

- $P_k$ := $\{x \in P | K[k-1] < x\}$.

Because we clamped $P$, $P_0$ and $P_k$ are empty. Therefore $\bigcup_{i=1}^{k-1} P_i = P$.

A \lstinline{PWLCurve} is linear between consecutive control points, so $C$ is linear over each $P_i$.

\subsubsection{Convert each partition into two points}

From the theorem, for each $P_i$, we can produce a two-point set $P_i'$ with $min_x(P_i) <= min_x(P_i') <= max_x(P_i') <= max_x(P_i)$, such that for any line $L$,

$SE(L, P_i) = SE(L, P_i') + SE(bestfitline(P_i), P_i)$. $C$ is linear over each $P_i$, so 

$SE(C, P_i) = SE(C, P_i') + SE(bestfitline(P_i), P_i)$.

\subsubsection{Recombine partitions}

Let $P':= \bigcup_{i=1}^{k-1} P_i'$. Each $P_i'$ consists of two points, so $P'$ consists of $2(|K|-1)$ points.

\begin{align*}
    SE(C, P) &= \sum_{i=1}^{k-1} SE(C, P_i) \\
    &= {\sum_{i=1}^{k-1} (SE(C, P_i') + SE(bestfitline(P_i), P_i))} \\
    &= {SE(C, P') + \sum_{i=1}^{k-1} SE(bestfitline(P_i), P_i)}.
\end{align*}

$\sum_{i=1}^{k-1} SE(bestfitline(P_i), P_i)$ is determined by $P$ and $K$, and is therefore the same for every $C$. Therefore we've proven the corollary.

\end{document}